\documentclass[a4paper,USenglish,cleveref, autoref, thm-restate,authorcolumns]{oasics-v2021}
\usepackage{lineno}
\nolinenumbers

\title{TURBO: Utility-Aware Bandwidth Allocation for Cloud-Augmented Autonomous Control}
\titlerunning{TURBO: Real-time Utility-Aware Bandwidth Allocation}

\author{Peter Schafhalter$^{*}$}{University of California, Berkeley}{pschafhalter@berkeley.edu}{https://orcid.org/0009-0007-9865-0456}{}
\author{Alexander Krentsel$^{*}$}{University of California, Berkeley}{akrentsel@berkeley.edu}{https://orcid.org/0009-0005-2728-1898}{}
\author{Hongbo Wei}{University of California, Berkeley}{hwei0@berkeley.edu}{https://orcid.org/0009-0001-4824-5766}{}
\author{Joseph E. Gonzalez}{University of California, Berkeley}{jegonzal@berkeley.edu}{https://orcid.org/0000-0003-2921-956X}{}
\author{Sylvia Ratnasamy}{University of California, Berkeley}{sylvia@cs.berkeley.edu}{https://orcid.org/0000-0002-0524-9425}{}
\author{Scott Shenker}{University of California, Berkeley}{shenker@berkeley.edu}{https://orcid.org/0000-0002-1357-7533}{}
\author{Ion Stoica}{University of California, Berkeley}{istoica@berkeley.edu}{https://orcid.org/0000-0002-5373-0088}{}

\authorrunning{P. Schafhalter and A. Krentsel et al.}

\Copyright{Peter Schafhalter, Alexander Krentsel, Hongbo Wei, Joseph Gonzalez, Sylvia Ratnasamy, Scott Shenker, and Ion Stoica}



\usepackage{graphicx}
\usepackage{amssymb}
\usepackage[T1]{fontenc}
\usepackage[utf8]{inputenc} % allow utf-8 input
\usepackage[T1]{fontenc}    % use 8-bit T1 fonts

\usepackage{microtype}
\usepackage{graphicx}
\usepackage{xspace}
\usepackage{booktabs} % for professional tables
\usepackage{subcaption}
\usepackage{siunitx}
\usepackage{multirow}

\usepackage{gensymb}

\usepackage{amsmath}
\usepackage{amssymb}
\usepackage{mathtools}
\usepackage{amsthm}
\usepackage{soul}
\usepackage{xcolor}
\usepackage{nicefrac}       % compact symbols for 1/2, etc.
\usepackage{microtype}      % microtypography
\usepackage{makecell}
\usepackage{float}
\usepackage{dsfont} % indidator functions.

\usepackage{hyperref}
\hypersetup{
    colorlinks,
    linkcolor={red!50!black},
    citecolor={blue!50!black},
    urlcolor={blue!80!black},
    breaklinks=true,
}
\usepackage{xurl}

\newcommand{\eg}{{\it e.g.,}\xspace}
\newcommand{\ie}{{\it i.e.,}\xspace}

\newcommand{\sysnamelong}{Task Utility and Resource Bandwidth Optimizer\xspace}
\newcommand{\sysname}{TURBO\xspace}

\usepackage[style=base,textfont={small,it},belowskip=0pt]{caption}

\newcommand{\myparagraph}[1]{\noindent{\bfseries #1}}

\newcommand{\topic}[1]{\textcolor{orange}{\bf #1}}
\renewcommand{\topic}[1]{#1}

\newcommand{\cameraready}[1]{\noindent{\color{red} #1}}
\renewcommand{\cameraready}[1]{#1}

\EventEditors{Katerina J. Argyraki and Aurojit Panda}
\EventNoEds{2}
\EventLongTitle{1st New Ideas in Networked Systems (NINeS 2026)}
\EventShortTitle{NINeS 2026}
\EventAcronym{NINeS}
\EventYear{2026}
\EventDate{February 10, 2026}
\EventLocation{Virtual Conference}
\EventLogo{}
\SeriesVolume{139}
\ArticleNo{18}

\begin{document}
\maketitle

\noindent$^{*}$These authors contributed equally to this work.

\begin{abstract}

  Autonomous driving system progress has been driven by improvements in machine learning (ML) models, whose computational demands now exceed what edge devices alone can provide. The cloud offers abundant compute, but the network has long been treated as an unreliable bottleneck rather than a co-equal part of the autonomous vehicle control loop. We argue that this separation is no longer tenable: safety-critical autonomy requires co-design of control, models, and network resource allocation itself.

  We introduce \sysname, a cloud-augmented control framework that addresses this challenge, formulating bandwidth allocation and control pipeline configuration across both the car and cloud as a joint optimization problem. \sysname maximizes benefit to the car while guaranteeing safety in the face of highly variable network conditions. We implement \sysname and evaluate it in both simulation and real-world deployment, showing it can improve average accuracy by up to 15.6\%pt over existing on-vehicle-only pipelines. Our code is made available at \url{www.github.com/NetSys/turbo}.

  % Recent progress in machine learning (ML) has driven key advances in many real-time intelligent systems, including autonomous driving, warehouse robotics, delivery drones, and virtual reality. However, growth rates in ML model sizes (i.e. FLOPs, parameter counts) are outpacing the compute available on-board these devices. In this work, we trace these trends out for the grounding example of autonomous vehicles (AVs), and find that the growing discrepancy between compute requirements and resource constraints in end devices will benefit increasingly from cloud augmentation.
  % % 
  
  % In order to make cloud augmentation feasible, we argue that such control systems \textit{must} be explicitly co-designed with network control in mind. We discuss the implications of this requirement, and present our formulation for optimally allocating network resources to maximize control system utility in the face of variable network conditions. We further discuss open questions and applicability beyond AVs.
\end{abstract}

\section{Introduction}
\label{s:introduction}

\topic{
Autonomous driving holds huge transformative potential for society, leading the first wave of real-world machine learning (ML) system applications.} Autonomous vehicles (AVs) have the potential to reduce
road fatalities through the elimination of human
error~\cite{nhtsa-sae-automation},
free up to one billion hours spent in traffic per day by improving
traffic flow~\cite{mckinsey-50mins}, 
and provide mobility to millions of people impacted by
disabilities~\cite{claypool2017self}.
Recent years have seen successful limited commercial deployments of AVs~\cite{waymo-scaling-to-four-cities, cruise-austin, tesla-robotaxi} in target markets with favorable environments. However, challenges remain such as operation in poor weather conditions, construction zones, and busy regions~\cite{bloomberg-self-driving-is-going-nowhere}.
%

% \topic{AV deployment progress has been enabled by ML model advances in the components that make up the AV control pipeline.} This pipeline is structured as a DAG of ML-based modules, each responsible for tasks such as camera stream object detection, movement prediction, and action planning (\cref{s:background}).    
\topic{
Progress in autonomous vehicle deployment has been driven by advances in machine learning models across components of the AV control pipeline.
}
This pipeline is structured as a \cameraready{directed acyclic graph (DAG)} of ML-based services, each responsible for tasks such as camera stream object detection, movement prediction, and action planning (\cref{s:background}).    
Significant effort has been devoted to increasing the accuracy of the machine learning (ML)
models~\cite{chauffeurnet,wu2023point,shi2024mtr++,leng2024pvtransformer,mu2024most} that implement these services, which in turn improves the end-to-end decision-making of
AVs~\cite{ntsa-uber-collision,nhtsa-cruise-incident-2023}.
Recently, state-of-the-art (SOTA) models have shown remarkable improvements in
accuracy by scaling to larger parameter
counts~\cite{sevilla2022compute,zhai2022scaling}.

\begin{figure}
  \centering
  \includegraphics[width=0.5\columnwidth]{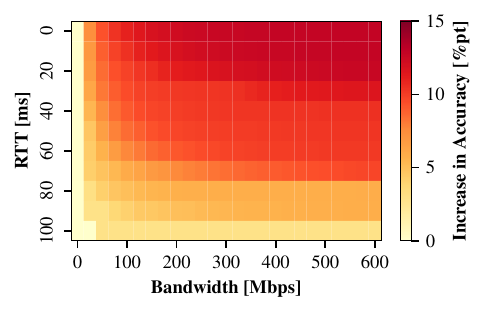}
  \caption{\sysname improves the average accuracy of AV perception and motion prediction services across a range of network conditions.}
  \label{f:heatmap}
\end{figure}

\topic{However, deploying SOTA models on AVs is increasingly challenging because vehicles \cameraready{have} an order-of-magnitude less compute than a single datacenter GPU\footnote{A
single SOTA cloud GPU (H100) can perform over $10\times$ more operations
per second than AV-targeted chips like NVIDIA's DRIVE
Orin~\cite{h100-spec,drive-orin-spec}.}.}
On-car compute is fundamentally limited by physical power, thermal, and stability
limits~\cite{driving-michigan,waymo-6th-gen-driver}, as well as the
high cost of ML accelerators that makes scaling on-vehicle compute financially infeasible. The tight runtime service-level-objectives (SLOs) for AV control tasks -- aiming to operate end-to-end with faster-than-human reaction times
(\eg $0.39$ to $1.2$ seconds~\cite{wolfe2020rapid,johansson1971drivers}) -- require system designers to carefully balance on-car model runtime with high-quality decision-making~\cite{gog2021pylot,tan20efficientdet,waymo-open-motion}.

% \topic{Given that the disparity between edge and cloud compute only continues to grow, we ask: how can we design a real-time edge control system that enables running larger, more accurate models than possible on the car, while maintaining control safety?} Here, control safety refers to \textit{guaranteeing} control decisions completing under any network conditions, including full disconnection. Our goal is to maximize control system performance (\ie accuracy of individual control modules) using the network bandwidth that \textit{is} available to the system, while maintaining this control safety.

\topic{Given the growing disparity between edge and cloud compute, we ask: how can we design a
real-time edge control system capable of running highly accurate, compute-intensive models
without compromising safety?}
The key safety requirement we consider is the ability to make decisions within a real-time constraint in any operating environment.
Because we aim to bridge this disparity using mobile networks, our system must continuously meet this requirement under unreliable network conditions as well as full disconnection.
Under this requirement, we design our system to maximize the decision-making quality,
\ie the accuracy of the individual services comprising the AV pipeline.

% \topic{The principal challenge for such a system is that running the control pipeline off the car requires traversing cellular networks, which can have severely limited and highly variable network conditions compared to traditional networked environments.} 
\topic{Meeting these design goals in a practical system presents several challenges.} The primary challenge is that running the control pipeline off the car requires traversing cellular networks, which have severely limited and highly variable network conditions compared to traditional networked environments. AVs generate over 8 Gbps of data across different sensing modalities\footnote{A camera generating $1920\times1280$ frames, $10$ Hz contributes 590 Mbps.}~\cite{cruise-roscon,siemens-av-data}, far exceeding the $100$ Mbps target uplink bandwidth for 5G networks. Prior work~\cite{schafhalter2023leveraging} has observed that cloud-grade GPUs can run an individual ML model faster than on-car GPUs, fast enough to make up for cellular \textit{ping} \cameraready{round-trip time (RTT)} latencies\cameraready{;
however, this work does not discuss the crucial portion of latency induced by limited cellular bandwidths,
which is significant enough to make executing in the cloud infeasible as shown in \cref{sec:eval-performance}.
% however, this work crucially ignores the portion of latency induced by limited cellular bandwidths,
%which we show in \cref{sec:eval-performance} is significant enough to cloud use infeasible as presented.
}
%, but neglects addressing this bandwidth-induced delay that we identify as the key constrained resource.
Second, we make the key observation that the value of each data stream generated by an AV is not equal, and can vary dynamically with the environment.

\topic{Our key insight to address these challenges is the opportunity and need to \textit{co-design} the control pipeline with resource allocation.} The opportunity arises from the fact that unlike traditional coded systems, services in a compound AI control system are \textit{reconfigurable} through the availability of ``families'' of task-specific models. Each model in the family is interchangeable, providing its own runtime/size/accuracy tradeoffs for a given task. 
This allows our system to unify resource allocation and dynamic system configuration; \eg the system can increase bandwidth allocation to a datastream to lower transmission time and enable running a slower, more accurate model where it is holistically most beneficial.
The need arises as a consequence of differing value of datastreams; as we show later on (\cref{sec:eval-performance}), without intelligently selecting the datastreams to allocate bandwidth to, the car sees practically no performance benefit under real-world network conditions.

% Unlike the traditional resource allocation setting, here our formulation reconfigures the system implementation jointly with the resource allocation – that is, we co-solve the optimal compression schemes and \textit{bandwidth} allocations to datastreams on the vehicle, along with selecting the concrete module "implementation" to run as part of the control pipeline in the cloud. 

\topic{We present \sysname, a cross-edge-cloud AV control framework that jointly optimizes bandwidth allocation and cloud model selection to maximize benefit to the car.} \sysname – \sysnamelong – offers multiple cloud models for several AV services, and greedily selects the best possible configuration of both cloud models \textit{and} bandwidth allocations that will meet strict runtime SLOs, while continuing to run on-vehicle models in parallel as a fallback to ensure baseline safety. These decisions happen dynamically at runtime, allowing \sysname to maximize overall accuracy across network conditions and driving environments. \sysname does this by extending the concept of \textit{utility curves}~\cite{zegura1999utility,wang2006utilities} to capture the intra-application relative benefit of a particular allocations of bandwidth across the AV system, and formulating an ILP over aggregated utility curves to select the best set of allocations.

\topic{We evaluate \sysname by running it live on a car deployment on highway and neighborhood streets, and through testing in simulation across hundreds of hours of real-world AV traces collected by Waymo~\cite{waymo-open-dataset}.} Our results show that when operating with SOTA open-source models (\cref{s:design}) on a real-world AV dataset, \sysname improves average accuracy by up to
% Modified:
15.6 \cameraready{percentage points} (\%pt)
over executing on-vehicle only and 12.7 \%pt over naive bandwidth allocation methods (\cref{s:evaluation}). Our real-world test drive shows our approach is feasible to run even with cellular network conditions today. We release our system code at \url{www.github.com/NetSys/turbo}.

\section{Background}
\label{s:background}

\subsection{Anatomy of an Autonomous Vehicle}
\label{s:background:av}
AVs capture information about their surroundings using sensors and process
that sensor data into control commands (\ie steering, acceleration, and
braking).
Data processing must be timely
and accurate,
presenting a critical challenge to the development of autonomous driving.
The computation in an AV is typically structured as a pipeline where each
component performs a specific task (\eg detecting nearby
obstacles)~\cite{gog2021pylot,cruise-roscon}.
To ensure the timely computation of control commands, pipelines execute
under an end-to-end deadline~\cite{erdos} with the potential to outperform
human reaction times of $390$ milliseconds to $1.2$
seconds~\cite{wolfe2020rapid,johansson1971drivers}.

To retrieve meaningful information from sensor data and enable intelligent
decision-making, AVs employ ML models. %  for several components.
% The quality of ML models is evaluated based on the accuracy
ML models are evaluated on
on large, offline datasets~\cite{coco,kitti-detection,waymo-open-dataset,caesar_nuscenes_2020,argo-data} where key innovations often result in low single-digit, but statistically significant, percentage point increases in accuracy~\cite{zong2023detrs,zhang2022dino,vit,liu2021swin} that indicate new capabilities.
Because more accurate ML models are generally more
compute-intensive~\cite{sevilla2022compute,zhai2022scaling,tan20efficientdet},
constructing an AV pipeline requires careful consideration of the tradeoffs
between accuracy and response time.
We identify that \textit{AV components are services} because their tasks
define concrete interfaces and their implementations select from a diverse
set of potential models and algorithms.
We provide an overview of the key components of an AV pipeline, examine
how each component forms a service, and discuss the role of ML.
%

% \vspace{-1em}

\subsubsection{Sensors}
\label{s:background:av:sensors}
AVs use high-fidelity sensors which span several modalities to observe
their surroundings.
%A collection of
\textit{Cameras} captures images from multiple % several different
perspectives.
\textit{Lidars} generate point clouds by using a rotating array of laser beams to measure the distance to nearby obstacles.
\textit{Radars} measure the distance, direction, and velocity of nearby
obstacles by emitting radio waves and measuring their reflections.
\textit{External Audio Receivers} capture sounds to detect and localize
emergency vehicles sirens.

Taken together, AVs sensors capture a large amount of detailed information
in order to generate a $360\degree$ view of the vehicle's environment which
aims to be accurate
at distances up to $500$ meters~\cite{waymo-6th-gen-driver}.
To increase fidelity and provide redundancy, AVs are equipped with multiple
instances of each sensor type. % (\cref{t:av-sensors}).
For example, Waymo's 5\textsuperscript{th} generation AV uses a long-range
camera to detect distant obstacles and peripheral cameras to reduce blind
spots~\cite{waymo-av-sensors}.
While sensor configurations vary, % and impact the amount of data generated,
open-source driving datasets indicate that a single camera may generate between 479 Mbps\footnote{Waymo provides 1920$\times$1040 images sampled at $10$ Hz~\cite{waymo-open-dataset}.}
and 1.8 Gbps.\footnote{The Argoverse dataset~\cite{argo-data} uses 1920$\times$1200 cameras recording at 33 Hz.} 

% \vspace{-1em}
\subsubsection{Perception}
\label{s:background:av:perception}
To understand their surroundings, AVs use ML models to
process sensor data into an ego-centric map of nearby obstacles, driveable
regions, and traffic annotations (\eg signs, traffic lights).
Perception performs several different tasks such as object detection,
object tracking, and lane detection~\cite{gog2021pylot,apollo-baidu} which
form subservices that may process data from different sensors.
While there is a large range of perception models for autonomous driving
which process different sensor modalities, most of these models use
convolutional neural networks (CNNs) to extract features from
images or lidar point
clouds~\cite{karpathy-keynote-cvpr-wad-2021,yin2021center,tang-lane-detection-review}.

In this work, we examine how to allocate bandwidth across 2D object
detection services, where detection on each camera stream forms a
distinct service.
Object detection is a well-studied perception task with a wide variety of
open-source models~\cite{tan20efficientdet,carion2020end,resnet,yolov4} and
performs the safety-critical task of identifying and locating nearby
obstacles by processing images from the AV's cameras.

Object detection models generate labeled bounding boxes that identify the
positions and the classes (\ie types) of objects in an image
(\cref{f:ed1-vs-ed7x}), and are evaluated using the following metrics.
\textit{Average Precision} (AP) measures whether a model's predicted
bounding boxes overlap with the true bounding boxes of the same class and
penalizes the model for false positives.
\textit{Mean Average Precision} (mAP) reports the average AP across all
object classes~\cite{pyimagesearch2022cocomap}.
In this paper, we use mAP to describe the accuracy of object detection
models because it is the standard detection accuracy metric across datasets
and
leaderboards~\cite{waymo-open-dataset,huggingface-object-detection-leaderboard}.

% \vspace{-1em}
\subsubsection{Motion Prediction}
\label{s:background:av:prediction}
Motion prediction uses ML models to anticipate the motion of nearby agents
(\eg pedestrians, vehicles, bicyclists) by processing the outputs from
perception.
State-of-the-art prediction % \alex{should we say "state-of-the-art" here?}
models typically leverage compute-intensive neural networks such
as Transformers~\cite{vaswani2017attention} to forecast the future positions of nearby agents based on
patterns in their behaviors and motion~\cite{rhinehart2018r2p2,shi2022motion,ettinger2024scaling};
however, some pipelines use simple compute-efficient models such as linear
regression~\cite{gog2021pylot}.

While there are several different metrics to evaluate motion prediction
models, we likewise use mAP as described in
\cite{waymo-motion-prediction-challenge}.
For motion prediction, mAP represents agents as bounding boxes on a 2D
top-down map, and uses the overlap between the predicted and true
positions to estimate the accuracy at a particular point in time.
When reporting the mAP, this accuracy is averaged across agent classes and
time steps.

\subsubsection{Remaining Components}
\label{s:background:av:other-components}

AVs further rely on \textit{localization}~\cite{nvidia-localization} to estimate the vehicle's position,
\textit{planning} to generate safe and reliable motion plans~\cite{katrakazas,paden-survey,hu2023planning,teng2023planning-survey,lu2023imitation},
and \textit{control} to convert motion plans into steering, acceleration, and braking commands~\cite{guanetti_control_2018}.
While localization provides motion prediction with access to high-fidelity maps, planning and control are downstream components that rely on accurate results from perception and motion prediction to make driving decisions.

\subsection{Remote Interventions}
\label{s:background:remote-interventions}

AVs already rely on cellular networks for safety-critical decision-making
for remote interventions.
When in an uncertain situation (\eg construction zones), the
AV contacts a remote human operator for
guidance~\cite{nytimes2024selfdrivinghelp, waymo-fleet-response}.
Using transmitted sensor feeds and annotated representations of the AV's surroundings,
the operator makes informed decisions to help the AV proceed \eg by creating a route for the AV to follow or answering clarifying questions posed by the AV.

% answering a series of questions posed by the AV to help it make a safe driving decision.
% \joey{I would want to know more about this?  Do they send all the video feeds?  Isn't the vehicle remotely driven?}
% Recent work has focused on the potential of cloud computing 
% Notably, recent work demonstrates that using the cloud opportunistically
% to conduct safety-critical computation can boost accuracy and has the
% potential to improve safety over only on-vehicle
% computation~\cite{schafhalter2023leveraging}.
% \peter{Redundant since we also mention this in the intro and the
% motivation?}
% %

\section{Motivation}
\label{s:motivation}

Highly accurate ML
models are key to ensuring that AVs
can make safe and reliable decisions.
These models enable processing huge volumes of sensor data into a detailed representation of an AV's surroundings, as well as understanding and forecasting the
behaviors of nearby human agents.
However, technical and economic trends challenge the ability to deploy the highest-accuracy models.

\begin{figure}
  \centering
  \includegraphics[width=0.7\columnwidth]{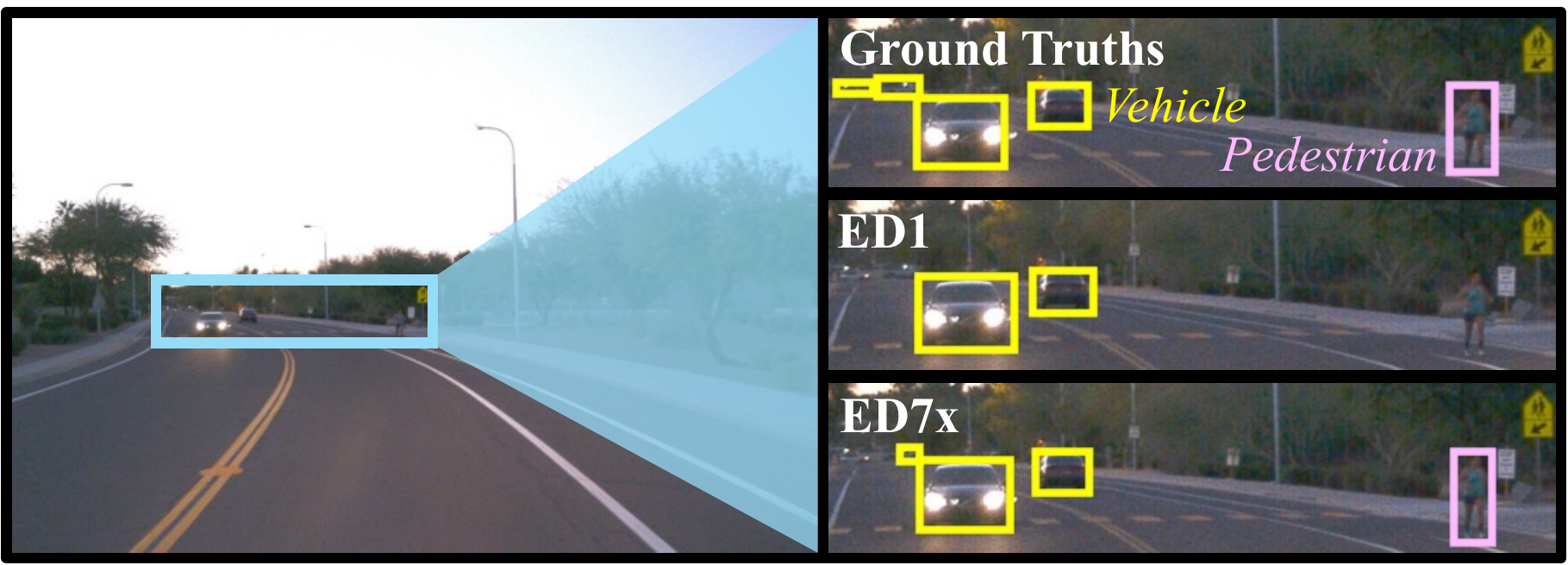}
  \caption{\textbf{More accurate detectors provide a better understanding
  of the surroundings.} In this scene from the Waymo Open
  Dataset~\cite{waymo-open-dataset}, EfficientDet-D1 (ED1 -- center)
  detects the two nearby vehicles but misses the pedestrian on the side of
  the road, resulting in a mean average precision (mAP) of 25\%.
  In contrast, highly accurate ED7x model (bottom) detects the pedestrian and
  both nearby vehicles, but generates an incorrect bounding box for one of the distant vehicles, resulting in an mAP of 75\%.}
  \label{f:ed1-vs-ed7x}
\end{figure}

The scalable ML model architectures ~\cite{sevilla2022compute,zhai2022scaling,liu2021swin} (\eg
Transformers~\cite{vaswani2017attention})
% that have enabled recent
that underpin state-of-the-art models for tasks pertinent to driving applications~\cite{yolov6,carion2020end,wayformer}
have also increased the computational resource requirements for inference by greatly increasing the total number of model parameters.
For this reason, ML practitioners develop \textit{families} of models with different tradeoffs between the number of parameters and the accuracy of a model (\cref{f:ed1-vs-ed7x}).
%
% The most accurate ML models are typically the most computationally
% expensive due to advances in scalable ML model
% architectures~\cite{sevilla2022compute,zhai2022scaling,liu2021swin} (\eg
% Transformers~\cite{vaswani2017attention}) which enable state-of-the-art models for 
%  driving applications~\cite{yolov6,carion2020end,wayformer}.
%
% Such models exhibit the remarkable ability to generate higher quality and
% more accurate outputs as the number of trainable parameters increases.
% However, models with higher parameter counts generally require more
% computational resources for training and inference.
%
% Taking advantage of scalable architectures, ML practitioners typically
% develop \textit{families} of models with different tradeoffs between
% the number of parameters and the accuracy of a model
% (\cref{f:ed1-vs-ed7x}).
%
For example, the EfficientDet family of object detection
models~\cite{tan20efficientdet} uses a scalable convolutional neural
network architecture~\cite{tan2019efficientnet} to develop 9 different
models with varying parameter counts and accuracies.
The smallest model, EfficientDet-D0 contains 3.9M parameters, requires 2.5B
floating point operations (FLOPs) to process a single image, and achieves a
mean average precision (mAP) of 34.3\% on the COCO validation
dataset~\cite{coco}.
In contrast, the largest model, EfficientDet-D7x, contains 77M parameters,
requires 410B FLOPs to process a single image, and achieves a mAP of
54.4\%.

Unfortunately, on-vehicle compute is limited by a variety of technical and economic constraints. Prior work~\cite{driving-michigan} shows AV compute can
significantly degrade the driving range of electric AVs due to the energy
needed for computation and cooling, and recommends that AV hardware must
tolerate significant impulses and vibrations. Moreover, the cost of deploying SOTA ML accelerators on AVs is prohibitive; an NVIDIA H100 GPU costs \$30k-40k in
2024~\cite{2024-h100-price,h100-price-june-2024}, as much as a new Tesla
Model 3~\cite{tesla-model-3-price}, and cost is already a key factor in the design of AVs~\cite{waymo-6th-gen-driver}. Along with spatial constraints, these limitations prevent AVs from deploying powerful chips or scaling compute on-vehicle.

Therefore, AV manufacturers leverage cost-effective compute platforms
such as NVIDIA's DRIVE Orin which powers autonomous driving for Volvo and
SAIC~\cite{drive-orin-usage}, but performs over $10\times$ fewer operations
per second than the H100~\cite{h100-spec,drive-orin-spec}.
Due to these hardware limitations, AVs lack the processing power to execute
the most accurate models in real time.
Thus, AV developers must carefully navigate the tradeoffs between
runtime and accuracy at design time (\eg by selecting appropriate models and using
techniques such as quantization) to ensure that AVs can
make high-quality and safe driving decisions while providing rapid response
times~\cite{gog2021pylot,erdos}.

\begin{figure}
  \centering
  \includegraphics[width=0.5\columnwidth]{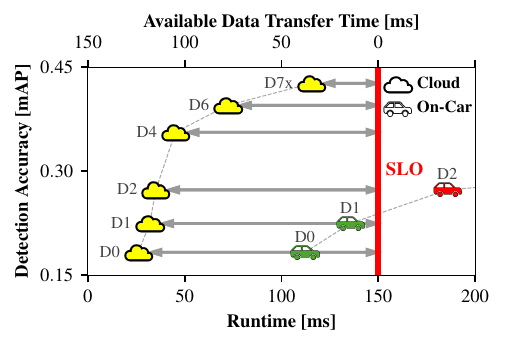}
  \caption{Cloud accelerators can execute more accurate models with lower
  runtime than hardware designed for autonomous driving, enabling AVs to
  increase accuracy while meeting stringent SLOs.
  % Here we use an NVIDIA H100 GPU as the cloud accelerator and a Jetson Orin to represent the on-vehicle hardware.
  }
  \label{f:latency-accuracy-tradeoff}
\end{figure}

Prior work in robotics observes that the cloud model runtime is fast enough to enable running more-accurate models within-SLO, while continuing to run an on-vehicle model as backup~\cite{schafhalter2023leveraging}.
However, their work does not focus on the effect of limited \textit{bandwidth}, only accounting for network latency.
As we show in (\cref{sec:eval-performance}), ignoring bandwidth limits accuracy improvement to under 2\%, as model inputs
are unable to reach the cloud in time to finish model execution by the SLO. % don't make it to the cloud in time to finish by the SLO.
The focus of this paper is on the impact of the network and specifically the role of intelligent bandwidth allocation on real-time edge control system accuracy.
We show that careful bandwidth allocation is crucial, unlocking up to 6$\times$ higher accuracy improvement. In addition, we present a complete end-to-end system implementation and evaluation.

\section{Method}
\label{s:method}

\newcommand{\tSLO}{t_\text{SLO}}
\newcommand{\inputSize}{S_\text{input}}
\newcommand{\tRTT}{t_\text{RTT}}
\newcommand{\tExec}{t_\text{exec}}
\newcommand{\modelsSet}{\mathcal{M}}
\newcommand{\servicesSet}{\mathcal{S}}
\newcommand{\R}{\mathbb{R}}
\newcommand{\1}{\mathbb{1}}

The goal of our method is to maximize benefit to the car's safety by choosing which services to run in the cloud. In order to do this, we must decide how much bandwidth to allocate to each service based on currently available bandwidth and ping latency, then use these settings to execute the best-performing model which satisfies the service-level objectives (SLOs). To achieve this, we turn to the idea of ``utility'' as introduced in previous work~\cite{zegura1999utility}, which defined utility as the incremental benefit an application gets from an additional unit of bandwidth, and observed that real-world applications may have highly-variable utility for each incremental unit of bandwidth.

We observe that the compound AI system structure~\cite{compound-ai-systems-blog} of the AV control pipeline introduced in \cref{s:background} decomposes cleanly into individual services that have their own distinct ``utility curves'' mapping bandwidth allocation to overall control accuracy impact.
% \peter{Merge with another paragraph.}
Thus our method has two parts; first, at system design time, we compute the utility of each available model as a function of allocated bandwidth (\cref{s:method:model-utility}),
then compose the utility functions of all available models to generate the utility
functions of each service (\cref{s:method:service-utility}). 
% and finally derive an application-level utility function (\cref{s:method:application-utility}) that merges the utilities of all services.\alex{nit: our application-level utility curve is *formed by* solving the ILP for each point in the set of network conditions} 
Second, at runtime, using these utility functions, we formulate and solve an Integer Linear Program (ILP) to find the optimal bandwidth allocations for the ping latency and amount of bandwidth available
(\cref{s:method:bandwidth-allocation}) that maximizes the application-level utility (\cref{s:method:application-utility}). We describe each part below.

\subsection{Model-Level Utility}
\label{s:method:model-utility}
% COMMENTED OUT FROM SUBMISSION:
% A model's utility must take into account the model's performance on its task (\ie \textit{accuracy}) as well as whether the model completes its SLOs (\ie \textit{latency}).
% For a model $m$ with accuracy $A$, latency SLO $t_\text{SLO}$, and the
% total runtime $T_\text{rt}(b)$ which is a function of allocated bandwidth
% $b$ and the round-trip time (RTT) $\tRTT$, the utility function is:
\cameraready{
A model's utility must take into account the model's performance on its task (\ie \textit{accuracy}) as well as whether the model meets its SLO (\ie \textit{latency}).
Given a model $m$ with accuracy $A_m$ which must execute within a latency SLO $\tSLO$, we
design the utility as a unit step function which provides a utility of $A_m$ if $\tSLO$ is met.
If $\tSLO$ is missed, we return a utility of $0$ because AVs must make decisions in real-time
and late results provide no value.
We observe that allocated bandwidth $b$ and round-trip time (RTT) $\tRTT$ affect the observed
runtime of a model $T(b,\tRTT)$, leading to the utility function:
}

\begin{equation}
  \label{eq:model-utility}
  U_m(b) =
  \begin{cases}
    A_m & T(b, \tRTT) \leq \tSLO \\
    0 & T(b, \tRTT) > \tSLO \\
  \end{cases}
\end{equation}

%This function forms
% COMMENTED OUT FROM SUBMISSION:
% $U_m(b)$ is
% a unit step function where utility is $0$ when the SLO
% is violated, and utility equals the model's accuracy when the SLO is met.
Because on-vehicle models are configured to meet latency SLOs and are unaffected by network transfer times, the utility curves of on-vehicle models simplify to:
% Because on-vehicle models have runtimes that are unaffected by bandwidth
% and are configured to meet latency SLOs, the utility curves of on-vehicle
% models simplify to:

\begin{equation}
  U_\text{on-vehicle model}(b) = A_\text{on-vehicle model}
\end{equation}

In contrast, the runtime of models executing on remote resources depends on
the allocated bandwidth, the characteristics of the model, and the network
conditions.
Given model input size $\inputSize$, round trip time $\tRTT$, and execution
time $\tExec$, the total runtime of the model is:

\begin{equation}
  T_\text{remote model}(b, \tRTT) = \tExec + \tRTT + \frac{\inputSize}{b}
\end{equation}

This produces the following utility curve which is a step function from $0$
to $A_\text{remote model}$ where the step occurs at $b_c$ when $T(b_c, \tRTT) = \tSLO$:

\begin{gather}
  U_\text{remote model}(b) = 
  \begin{cases}
    A_\text{remote model} & \tExec + \tRTT + \frac{\inputSize}{b} \leq \tSLO \\
    0 & \text{otherwise} % \tExec + \tRTT + \frac{\inputSize}{b} > \tSLO \\
  \end{cases} \\
  b_c = \frac{\inputSize}{\tSLO - \tExec - \tRTT}
  \label{eq:bandwidth-step}
\end{gather}

A key consequence of the step-utility function is that $b_c$ is the minimal
amount of bandwidth to run a remote model which is optimal when bandwidth
is scarce.
Bandwidth allocations greater than $b_c$ provide the same utility of $A_m$
as a bandwidth allocation of exactly $b_c$; consequentially, any excess
bandwidth is wasted.
Moreover, any bandwidth allocation less than $b_c$ has a utility of 0, so
bandwidth allocations $0 < b < b_c$ are also wasted. % \ak{good!}

We also observe that more accurate models typically have larger inputs
$\inputSize$ and longer execution times $\tExec$.
Consequently, more accurate models typically require larger bandwidth
allocations $b$ and are less tolerant of large round trip times $\tRTT$
compared to their less accurate counterparts.
% [DONE] \peter{Add a figure/table, maybe in motivation.}

\subsection{Service-Level Utility}
\label{s:method:service-utility}

A \textit{service} consists of a specific task, such as detecting obstacles on a particular camera stream, as well as several models that can perform that task with different parameters and resource requirements.
The service typically runs on-vehicle the best model that on-vehicle resources allow, and additional models can execute remotely on cloud-based hardware.
To ensure that the service tolerates sudden disconnections and
reductions in bandwidth, the service always executes the on-vehicle model
as suggested in \cite{schafhalter2023leveraging}.
In this way, \textbf{we guarantee that the service will always provide a utility of at least
$A_\text{local model}$}, ensuring that our approach of attempting to use remote models is
\textit{strictly better} than using only local models.

We define the ``utility function'' of a service $s$ as the amount of control-application utility gained by granting a certain amount of bandwidth $b$ to the service.

% COMMENTED OUT FROM SUMISSION:
% Formally, we define this as the maximum utility of all of the service's models $\modelsSet$:
\cameraready{
To ensure the service selects the most accurate model that satisfies the SLO,
we define the service-level utility as the maximum utility among all of the service's models $\modelsSet$:
}

\begin{equation}
  U_s(b) = \max_{m \in \modelsSet} U_m(b)
\end{equation}

By applying the model utility function defined by \cref{eq:model-utility}, we find that $U_s(b)$ equals the utility of the most accurate model that can satisfy its SLO with the provided bandwidth. 

\begin{figure}
    \centering
    \includegraphics[width=0.7\columnwidth]{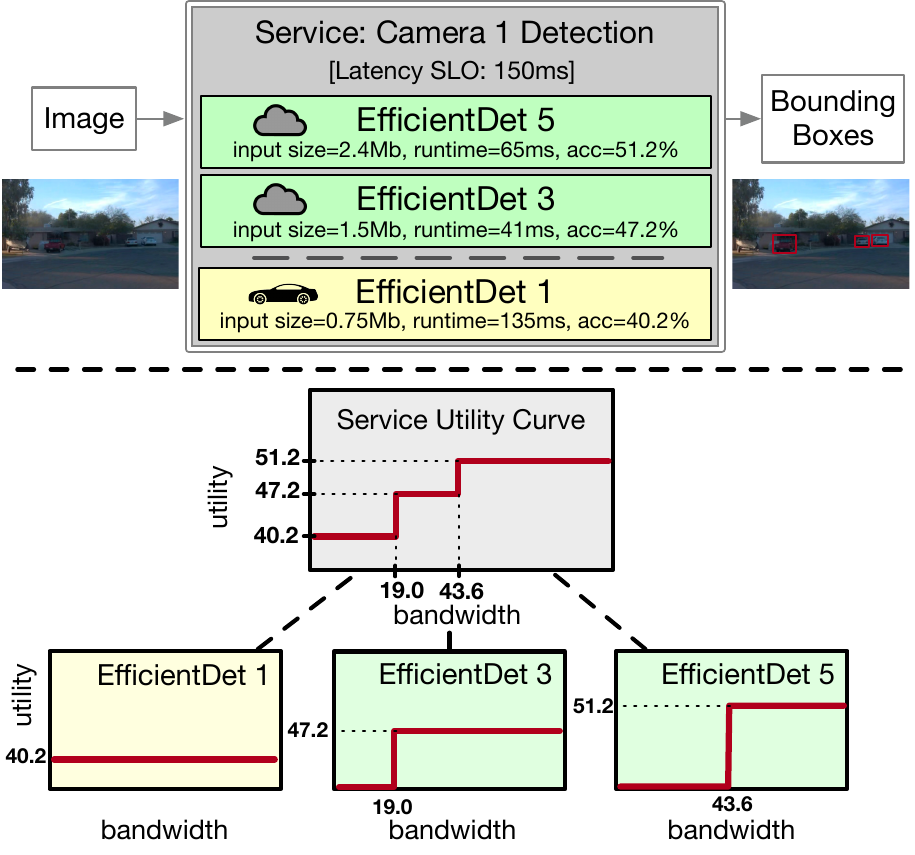}
    %\vspace{-0.5em}
    \caption{Utility curves for a service with 3 object detection models.
    }
    \label{f:utility-curves}
    %\vspace{-1.5em}
\end{figure}

We show an example of the construction of a service-level utility curve for a single object detection service which processes a video stream using a small EfficientDet D1 (ED1) model available on the car and two larger models, ED3 and ED5, available in the cloud, shown in \cref{f:utility-curves}. Because more accurate models require more bandwidth to meet their SLOs, the shape of $U_s(b)$ is a series of steps which occur wherever the service has sufficient bandwidth to transition to a more accurate model. Thus, the optimal bandwidth allocation for a service is the selected model's optimal bandwidth $b_c$, or 0 if running a local model, and any excess bandwidth is wasted.

Under our definition, each service executes at most 1 remote model\footnote{It may also be possible to run multiple cloud models and return results from each as they become available using the best results returned before the SLO, however this may become computationally cost prohibitive; we do not consider such a design here.}.
Therefore, setting a bandwidth allocation for a service effectively selects which model (or none) to run remotely. 
% \peter{Discuss compression for model inputs (\eg variable bit encoding).
% How compressible is the video?
% How does compression impact accuracy?
% ImageMagick: change compression level.
% }\alex{I think we ought to wave this off to say it's orthogonal, no? Compressing an input doesn't affect our method, you just add more time to model runtime, and decrease the size of the input?}

\subsection{Application-Level Utility}
\label{s:method:application-utility}

To achieve optimal performance on application level goals (\eg safety for
autonomous driving), we must coordinate bandwidth allocation across
services.
The key challenge lies in the fact that the relationship between
service-level utility and AV control system quality is not always well-defined
and may vary across different services; that is, overall service performance is not necessarily the average accuracy across all services, but rather the safety of the final end-result of how the car moves.

For example, detecting obstacles observed by the AV's front camera is
typically more important than detecting obstacles from the AV's side camera.
Complicating the matter, the performance of one service may impact the
performance of another given the DAG structure of compound AI systems, \eg motion prediction relies on an accurate history
of the motion of nearby obstacles, and the accuracy of this history is
impacted by the detection accuracy.

We propose a general framework which derives an overall application-level
utility by combining the utility functions of a set of services
$\servicesSet$:

\begin{equation}
  \label{eq:application-utility}
  U_\text{app}(b) = \max_{b_s : s \in \servicesSet}
  \sum_{s \in \servicesSet} f_s\big( U_s(b_s) \big)
  : \sum_{b_s : s \in \servicesSet} b_s \leq b
\end{equation}

We account for differences in the impact of each service's utility on the
application-level utility by transforming the service-level utility with
$f_s : \R \rightarrow \R$.
For example, setting $f_s(x) = ax + b$ can re-weight service-level
utilities to prioritize more important services.
The $f_s$ transformation may also be used to normalize the utilities of
different services so that they exist in the same range and can be more
easily compared (e.g., by incorporating a sigmoid function to convert
$U_s : \R^+ \rightarrow \R$ to $f_s \circ U_s : \R \rightarrow [0, 1]$.
We emphasize that $f_s$ should be chosen carefully, and evaluated or even
learned using real-world data or highly realistic simulations to ensure
that $U_\text{app}$ reflects the application's goals.
% that the $U_\text{app}$ reflects the application's goals.
% \peter{Suppose we have a simulated crash rate.
% Try many different utility functions.
% Learn a utility function that minimizes crash rate is future work.
% The focus of the paper isn't the design of the objective function, we show that our method generalizes to many different variants of objective functions ($f_s$).
% }

\subsection{Runtime Bandwidth Allocation}
\label{s:method:bandwidth-allocation}

Our goal at runtime is to decide which models to run in the cloud and thus
which input data to transfer to the cloud with the limited bandwidth
available.
In order to do this, we collect periodic estimates of available bandwidth
and RTT using standard methods~\cite{iperf3,nuttcp}.
% \alex{need to add bandwidth allocation enforcement mechanism?}
% Bandwidth and RTT latency has
% variance shown in \cref{f:to-be-added}, thus we sample every X seconds.\alex{remove?}

% In our problem, we aim to optimize the allocation of a limited bandwidth resource across multiple services, each with a discrete utility function defined by a series of intervals. The objective is to maximize the total utility obtained from all services. This problem is formulated as an Integer Linear Program (ILP), ...

We formulate the bandwidth allocation decision as a utility maximization
problem, given the available bandwidth, RTT, and the utility functions for
the models (\cref{s:method:model-utility}), services
(\cref{s:method:service-utility}), and application
(\cref{s:method:application-utility}).
Our formulation is an Integer Linear Program (ILP) that maximizes the
utility of the overall application, and leverages the fact that the utility
functions of models and services are step functions directly select which
configurations of cloud models to run.
% We thus formulate the control application-level allocation problem of
% maximizing bandwidth given in \cref{eq:application-utility} as an Integer
% Linear Programming (ILP) problem as follows where binary decision variables
% indicate whether a specific interval of utility is selected for each
% service, taking advantage of the fact that there are only a few, discrete
% service-level configurations (at the steps in the service-level utility
% curve).

Let $|\servicesSet|$ be the number of services, and note that steps of the
utility function for each service $s\in \servicesSet$ can be defined by the
accuracy $a_{s,m}$ and the location of the step $b_{c,s,m}$ for each model
configuration $m\in\modelsSet_s$ in the service.

We define binary decision variables:
\begin{equation}
x_{s,m} =
\begin{cases} 
1 & \text{if model } m \text{ of service } s \text{ is selected}, \\
0 & \text{otherwise}.
\end{cases}
\end{equation}

\noindent\textbf{Objective Function}: Maximize the total utility across all services:

\begin{equation}
  \max_{x}\quad \sum_{s\in\servicesSet} \sum_{m\in\modelsSet_s} x_{s,m}
  \cdot a_{s,m}
\end{equation}

\noindent\textbf{Constraints}:
Ensure that cumulative bandwidth allocated is below the available total,
and that only one or zero cloud-based model configurations are selected.

\textit{Bandwidth Constraint}:
The total allocated bandwidth across all services must not exceed the
available bandwidth:
\begin{equation}
  \sum_{s\in\servicesSet} \sum_{m\in\modelsSet_s} x_{s,m} \cdot b_{c,s,m} \leq B
  \label{eq:bandwidth-constraint}
\end{equation}
where $b_{c,s,m}$ is the bandwidth at which the step of the utility
function for model $m$ of service $s$ occurs.

\textit{One Cloud Model per Service Constraint}:
% We execute at most one model in the cloud for each service:
Each service must be allocated one model configuration:
% Each service can be allocated only one model configuration:
% \joey{Why only one model per service?  Could we say that, because each model has different input requirements or something?}

\begin{equation}
  \sum_{m\in\modelsSet_s} x_{s,m} = 1 \quad \text{for each } s \in
  \servicesSet
  \label{eq:one-model-constraint}
\end{equation}

\textit{Binary Constraint}:
The decision variables are binary:
\begin{equation}
  x_{m,s} \in \{0, 1\} \quad \text{for all } m \in \modelsSet_s, s \in
  \servicesSet
\end{equation}

The objective function maximizes the sum of utilities across all selected
intervals.
The bandwidth constraint in \cref{eq:bandwidth-constraint} ensures that the
total allocated bandwidth does not exceed the available bandwidth.
The one model per service constraint in \cref{eq:one-model-constraint}
enforces that each service selects one model configuration, which indicates that the solver should not consider solutions that run multiple models for a service in the cloud and waste bandwidth.
The decision variables are constrained to be binary, reflecting the
discrete nature of the utility allocation.

This ILP formulation directly selects which models configurations to run
for the cloud across a set of services while maximizing the overall utility
to the application.
Consequentially, solving the ILP also generates bandwidth allocations for
each service that are defined by the step $b_c$ of the selected models'
utility functions.
In \cref{s:design}, we apply this method to derive utility functions using
state-of-the-art models, and evaluate its ability to boost the accuracy of
autonomous driving services in \cref{s:evaluation}.

\section{Design and Implementation}
\label{s:design}

We design and implement the method described as a standalone control module. To evaluate its efficacy in a real-world environment, we additionally design and implement a edge-cloud offload control system runtime, which we evaluate in \cref{s:eval:performance-real-world}. We describe the design for the module and the wider system in turn below.

\subsection{Resource Model}
\label{s:design:resources}

% \begin{itemize}
%   \item Jetson Orin on-vehicle.
%   \item H100 in the cloud. Assume an H100 is always available and can begin
%     processing immediately (\eg by over-provisioning).
%     In-practice, we would use some kind of real-time serving system that
%     can also take advantage of economies of scale/batching to serve a
%     fleet efficiently and meet SLOs: cite tetrisched, clipper, etc.
% \end{itemize}

We model the AV as a resource-constrained edge device with an NVIDIA Jetson
Orin compute system because it uses the same chip as the NVIDIA DRIVE
Orin~\cite{nvidia-drive-faq} which powers autonomous driving for vehicles
from Volvo and SAIC~\cite{drive-orin-usage}.
% \alex{Peter, add a footnote here (or to the text) citing this solidly with the article we have in the slack and other citations. This seems to be hard to believe for our readers.}
%
We assume that the AV has sufficient compute resources to run a pipelined
AV software stack using resource-efficient ML models on-vehicle that meet a
minimum accuracy level and runtime SLO.
%, and could benefit from more accurate ML models
% which demand more compute.
%
We view the cloud as a resource-rich compute environment which can
immediately process incoming data with more accurate ML models.
For cloud hardware, we assume that models execute on an NVIDIA H100 GPU
from a RunPod $1\times$H100 PCIe instance with 176 GB
RAM and 16 vCPUs.
While there is significant work on serving systems for ML models, many
take advantage of economies of scale and batching to improve
efficiency~\cite{clipper,infaas,clockwork}.
We consider these systems to be complementary to our work as they
improve resource utilization and reduce costs when serving models at scale.

\subsection{Tasks}
\label{s:design:tasks}

\subsubsection{Object Detection}
\label{s:design:tasks:object-detection}
We first examine how object detection, a representative computer vision
task, can benefit from cloud computing.
Object detection is a well-studied task with a wide range of open-source
models~\cite{yolov4,ren2015faster,tan20efficientdet,carion2020end} and datasets~\cite{imagenet,coco,waymo-open-dataset,kitti-detection,bdd-100k} and uses model design patterns (\eg convolutional
neural networks) which are common in other perception tasks including
semantic segmentation and 3D object detection.
Our object detection SLO is $150$ ms\footnote{Object detection models may
meet tighter SLOs with model compilers, specialized runtimes, and model
compression techniques. We emphasize that our goal is to examine a
representative task for perception with a representative SLO in which cloud
computing permits the execution of more accurate and resource-intensive
models.}
which is similar to the perception runtimes of existing
systems~\cite{erdos,apollo-auto-github}.

\begin{table}
  \centering
  \begin{tabular}{|c||c||c|c||c|c|}
  \hline
\multirow{2}{*}{\textbf{Model}}
& \textbf{Input}
& \multicolumn{2}{c||}{\textbf{Preprocessing [ms]}}
& \multicolumn{2}{c|}{\textbf{Inference [ms]}} \\
  & \textbf{[Mb]} & \textbf{Orin} & \textbf{H100} & \textbf{Orin} & \textbf{H100} \\
\hline
  ED1 & 9.8 & 18 & 11 & 118 & 21 \\
\hline
  ED2 & 14.2 & 20 & 12 & 166 & 23 \\
\hline
  ED4 & 25.2 & 25 & 15 & 523 & 30 \\
\hline
  ED6 & 39.3 & 37 & 18 & 1350 & 54 \\
\hline
  ED7x & 56.6 & 43 & 25 & 2320 & 91 \\
\hline
\end{tabular}

  \caption{EfficientDet models. Preprocessing measures the runtime of
  resizing and preparing the image on CPU. Inference includes transferring
  pre-processed data to GPU and running the model.}
  \label{t:detection-models}
\end{table}

\begin{figure}
  \centering
  \includegraphics[width=0.6\columnwidth]{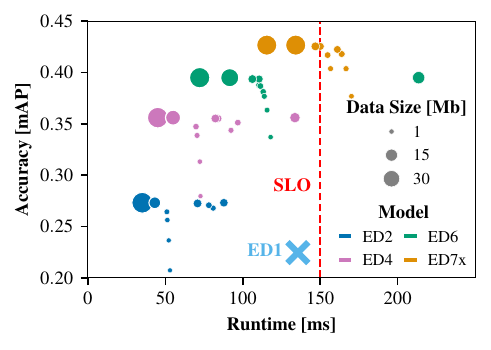}
  \vspace{-1em}
  \caption{We consider 35 model configurations with different tradeoffs in runtime, accuracy, and data transfer size:
  ED1 runs on-vehicle while
  ED2, ED4, ED6, and ED7x execute in the cloud. We generate additional
  configurations by applying lossless PNG and lossy JPEG compression to
  either the original image or the pre-processed model inputs.
  Our JPEG quality factors are 95, 90, 75, and 50.
  We find that compression significantly reduces the amount of data to
  transfer at the cost of increased runtime and reduced accuracy.}
  \label{f:cloud-model-properties}
  \vspace{-1em}
\end{figure}

\myparagraph{Models.} We study the EfficientDet family of
models~\cite{tan20efficientdet} because they provide a large trade-off
space between latency, accuracy, and resource requirements
(\cref{t:detection-models}).
We select \textbf{E}fficientDet-\textbf{D1} (ED1) as our on-vehicle model
because ED1 is the most accurate model that can meet the SLO using
on-vehicle hardware (\cref{f:latency-accuracy-tradeoff}).
Our cloud models are ED2, ED4, ED6, and ED7x, which are increasingly
accurate and resource-intensive and are all unable to meet the SLO using
on-vehicle hardware.
EfficientDet models include a lightweight on-CPU preprocessing step to
resize and prepare images for the deep neural network (DNN) running on the
GPU.
We exploit this preprocessing step as well as compression on images and the
along to generate the following configurations
for EfficientDet inference with different tradeoffs between runtime,
accuracy, and the amount of data to transfer
(\cref{f:cloud-model-properties}):
\begin{enumerate}
  \item \textit{Cloud preprocessing} transfers the original image to the
    cloud for preprocessing and neural network execution.
    This configuration shifts all processing to the cloud, resulting in the
    fastest runtime but the largest data transfer size.
  \item \textit{On-vehicle preprocessing} preprocesses the image on-vehicle
    and transfers the smaller inputs to the cloud. This configuration
    trades slower on-vehicle preprocessing for smaller data transfer.
  \item \textit{Image compression} compresses the original image using
    lossless PNG or lossy JPEG compression and transmits the compressed
    image to the cloud, where it is then decompressed.
    This configuration further reduces the amount of data transferred at the
    cost of higher runtimes due to compression and decompression.
  \item \textit{DNN input compression} preprocesses the image on-vehicle
    and similarly applies lossless PNG or lossy JPEG compression to the
    preprocessed model inputs which are then transmitted to the cloud.
    This configuration has the largest reduction in data size but the
    slowest (local) runtime.
\end{enumerate}

% %
% We further extend the cloud models by applying lossless PNG and lossy JPEG
% compression to either the preprocessed inputs or the original
% images\footnote{Compressing images moves preprocessing to the cloud.}.
% %
% We find that compression can significantly decrease the amount of data transferred
% at the cost of increased runtime and, for lossy compression, reduced
% accuracy (\cref{f:cloud-model-properties}).
% \peter{Discuss how we place and profile model runtime?}

% \begin{figure}
%   \centering
%   \includegraphics[width=0.9\columnwidth]{figures/cloud-detector-properties}
%   \caption{We select ED2, ED4, ED6, and ED7x as our cloud models and apply
%   compression to reduce the size of the
%   \textcolor[HTML]{3171AD}{pre-processed model inputs} or the
%   \textcolor[HTML]{469C76}{original input image}
%   using lossless PNG and lossy JPEG with quality factors of 95, 90, 75, and
%   50.
%   We find that compression can significantly reduce input size at the cost
%   of increased runtime and reduced accuracy.
%   }
%   \label{f:cloud-model-properties}
% \end{figure}

\myparagraph{Dataset.}
To train and evaluate detection models, we use the Waymo Open Dataset (WOD)
v2.0.0~\cite{waymo-open-dataset} which consists 6.4 hours of driving
organized into 1150 scenes of 20 seconds sampled at 10 Hz and contains 9.9
million annotated bounding boxes.
The WOD provides 5 camera perspectives which we view as distinct tasks.
The front, front-left, and front-right cameras generate images
at resolution of $1920 \times 1280$, while images from the side and
rear cameras are $1920 \times 886$, resulting in uncompressed data sizes of 
59 Mb and 41 Mb respectively.
The P99 number of ground truth bounding boxes per image is 62, translating
into an output size of 7.9 Kb.\footnote{Each bounding box comprises four
32-bit floating point numbers for the minimum and maximum x and y
values, and an 8-bit integer for the class.}

\myparagraph{Training.}
We partition the scenes from the WOD v2.0.0~\cite{waymo-open-dataset}
into training (68\%), validation (12\%), and test (20\%) sets.
To adapt the models to the dataset, we modify their classification heads to recognize five classes of objects: vehicles, pedestrians, cyclists, signs, and other.
We initialized the models with pre-trained weights from the COCO dataset~\cite{coco} provided by \cite{effdet-github}, and fine-tune on the WOD dataset for 10 epochs\footnote{We fine-tune ED7x for only 8 epochs due to the high
cost of training.}.

% Because WOD lacks public labels\alex{what are public labels? does WOD not provide us ground truthitself?} for its test set, we generate our own
% We generate
% training, validation, and test splits containing 68\%, 12\%, and 20\% of
% the WOD driving scenes respectively.
%
% We modify the models to categorize bounding boxes into 5 classes (vehicle,
% pedestrian, cyclist, sign, and other).
% %
% We initialize each model with pre-trained weights from the COCO
% dataset~\cite{coco} provided by \cite{effdet-github}, and fine-tune for 10
% epochs\footnote{We fine-tune ED7x for 8 epochs due to the high
% cost of training.} on the WOD.
% %
% \peter{TODO: discuss lower accuracy of ED models on Waymo. Attribute this
% to the difficulty of the dataset and differences in metric calculation.
% Add figure comparing 2 ED models vs. ground truths.
% }

\subsection{Motion Prediction}
\label{s:design:motion-prediction}
Motion prediction is a critical autonomous driving task which estimates the
future positions of nearby agents (\eg vehicles, pedestrians, and
cyclists).
Motion prediction is an active area of research with a variety of
approaches using different neural network architectures.
We select an SLO of $250$ ms for motion prediction based on the reported
runtimes of open-source AV implementations~\cite{erdos,apollo-auto-github}.
While the motion prediction models take both high-definition (HD) maps and
historical agent trajectories as inputs, we only transmit agent information
as maps can be pre-computed and stored in the cloud.

\myparagraph{Models.}
For the cloud model, we use Motion Transformer~\cite{shi2022motion},
a state-of-the-art model that employs a Transformer architecture to forecast trajectories and ranks first on the 2022 Waymo Open Motion Dataset\footnote{Waymo's Open Dataset and and Open Motion Dataset are distinct}~\cite{waymo-open-motion}
% \footnote{The Waymo Open
% Motion Dataset is distinct from the Waymo
% Open Dataset}
(WOMD) leaderboard.
For on-vehicle predictions, we select MotionCNN~\cite{konev2022motioncnn}, a lightweight model that ranked third on the 2021 WOMD leaderboard.
MotionCNN represents trajectories and surroundings as a fixed resolution image and uses a convolutional neural network to predict future paths.

% For the cloud model, we use Motion Transformer~\cite{shi2022motion} which
% ranks first on the 2022 leaderboard for the Waymo Open Motion
% Dataset (WOMD)~\cite{waymo-open-motion} and uses the transformer model
% architecture~\cite{vaswani2017attention} to predict future trajectories.
% %
% For the on-vehicle model, we select MotionCNN~\cite{konev2022motioncnn}
% which is a lightweight model that represents trajectories and nearby
% surroundings as a fixed-resolution image with many channels and applies a
% convolutional neural network to predict future trajectories.
% MotionCNN scored third on the 2021 Waymo motion prediction leaderboard.
% %

\myparagraph{Dataset.}
To train and evaluate the motion prediction models, we use WOMD
v1.2.1 which contains over 100k scenes of
20 seconds sampled at 10 Hz split into a training (70\%), validation
(15\%), and test (15\%) sets.
The WOMD includes 3D bounding boxes for each agent and map data
(\eg lanes, signs, crosswalks) for each scene.
Models must predict the positions of selected agents at three, five,
and eight seconds in the future using an HD map and a one second trajectory
history of each nearby agent.
Thus, the Motion Transformer model running in the cloud has a P99
input size of 146 Kb and a P99 output size of 246 Kb.
% \peter{Re-read for clarity.}
% The dataset provides one second of agent history and 

\myparagraph{Training.}
We modify the open-source implementations of Motion Transformer and MotionCNN for compatibility with WOMD v1.2.1, and follow the published training procedures. % to train the models.
Our trained models report a validation accuracy of 0.22 mAP for MotionCNN
and 0.40 for Motion Transformer.
% We use the open-source implementations to follow the published
% training procedures for Motion Transformer and MotionCNN.
% %
% We further contribute small adjustments to the training scripts for
% compatibility with WOMD v1.2.1.

\subsection{Utility Functions}
\label{s:design:utility}

We follow \cref{s:method:model-utility} to design utility functions for
each model configuration.
Our SLOs are $150$ ms for object detection and $250$ ms for motion
prediction.
We profile each cloud model using the Jetson Orin for pre-processing
and compression and the H100 for pre-processing, inference, and
post-processing and compute $\tExec$ using the P99 values.
We also measure the amount of data to transfer across the network for
each inference iteration and likewise set $\inputSize$ to the P99 value.
While images and pre-processed EfficientDet model inputs have constant
sizes, their compressed sizes vary (\cref{f:cloud-model-properties}).
% \peter{Include a figure or report statistics.}
%
Likewise, the Motion Transformer's input size varies with the number of
agents in the scene\footnote{We only transmit agent information because
map data can be pre-computed and stored.}.
Based on these profiles, we use \cref{eq:bandwidth-step} to calculate the bandwidth at which the step occurs $b_c$ which, along with the
accuracy, characterizes the model configuration's utility function.

To calculate the service-level utility, we include all cloud models as well
as the on-vehicle model.
Because the on-vehicle model does not transmit data, it provides a floor to
the performance of the service.
We further set $f_s(u) = u$ when calculating the application-level utility.
%
%Consequentially,
\cameraready{Therefore,} the application-level utility equals the average accuracy
across all services.

\subsection{Control Module Implementation}
\label{s:design:ilp}
We model our ILP in Python using PuLP~\cite{pulp-modeler} using the
measured accuracy, runtime, and data transfer requirements measured in
\cref{s:design:tasks} in conjunction with the RTT and available bandwidth according to \cref{s:method:bandwidth-allocation}.
We use the CBC solver~\cite{cbc-solver-2.10.12} to solve the ILP which
selects the cloud model configurations the object detection and motion
planning services.

\subsection{Edge-Cloud Offload Runtime Implementation}
\label{s:design:runtime-impl}

We demonstrate the real-world feasibility of AV bandwidth allocation by providing an end-to-end system implementation
involving
%\cameraready{with}
real-time network monitoring, edge cloud offloading, bandwidth allocation/multiplexing, and live camera feed ingestion. We design our
%implementation
\cameraready{system}
around the following tenets: to 1) minimize the latency incurred throughout the system; 2) ensure that a local on-car inference result is always available as a fallback if an offload request to the edge-cloud times out of the runtime SLO; and, 3) ensure that the bandwidth control module uses real-time monitors of network RTT and bandwidth conditions.

\begin{figure}[t]
    \centering
    \includegraphics[width=0.9\linewidth]{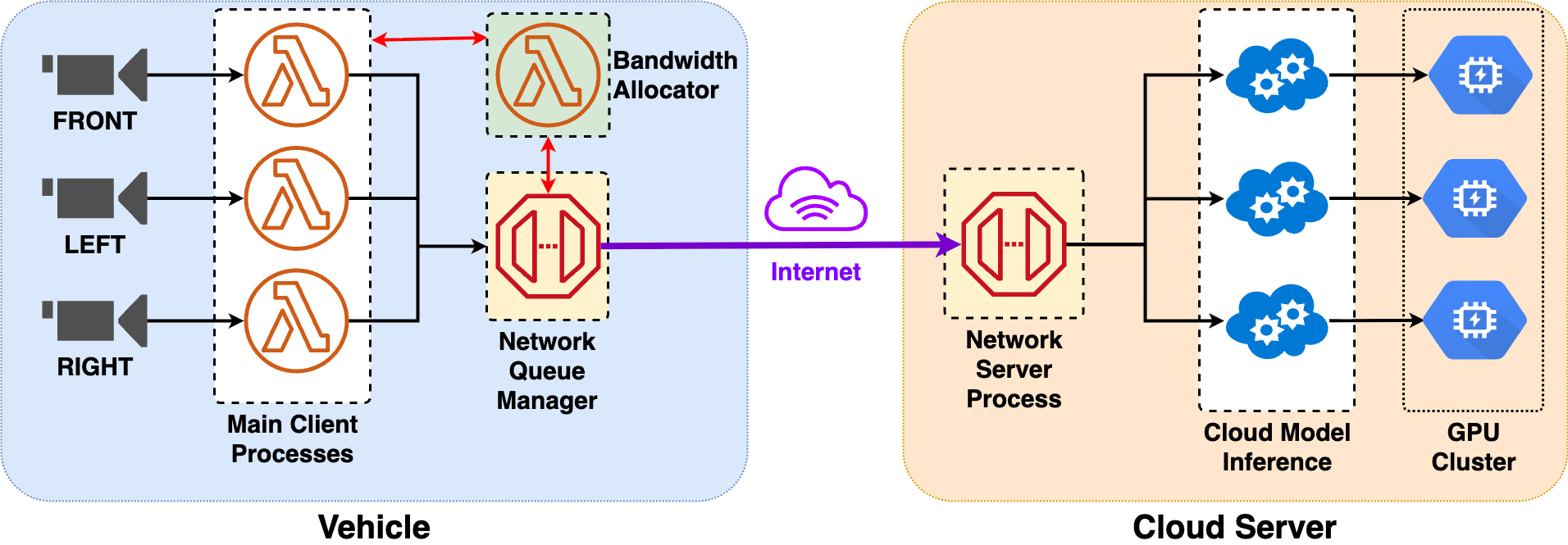}
    \caption{A high-level overview of the system implementation.}
    \label{fig:system-design-chart}
    \vspace{-1em}
\end{figure}

At a high level, our system implementation (\cref{fig:system-design-chart}) consists of two main parts: an on-vehicle client and the cloud server. The on-vehicle portion consists of a collection of on-car sensors (e.g. cameras);
%three 1080p webcams; 
a main client process that performs preprocessing and compression on the read camera images and writes them to the network queue process; a network queue manager that transmits images, receives cloud responses, enforces bandwidth constraints via queuing for each of the control services, and monitors network CWND and RTT conditions; and, a bandwidth allocator that runs our ILP optimization formulation based on the monitored network conditions. The cloud server portion consists of a network server process that receives and transmits from the on-vehicle network process, and PyTorch processes that run preprocessing (if necessary) and inference on received camera images. The system totals 1570 lines of Rust, and 3052 lines of Python.

For the network process that manages transfers between the on-vehicle and server portions of the system, we elect to use \texttt{s2n-quic}, a highly-optimized industry implementation of the QUIC protocol~\cite{awss2n-quic_2025}. We choose to use a QUIC implementation as our transport medium because the QUIC protocol~\cite{rfc9000} has two main features essential in our application: it (1) supports concurrent traffic streams on the same QUIC connection without head-of-line blocking, and (2) provides key live metrics about network conditions such as the RTT and CWND, which are necessary inputs to our bandwidth allocator control module. These features are enabled by QUIC running directly over UDP and bypassing the OS TCP stack, allowing for greater transparency about network conditions and fine grained control over traffic streams. 

\section{Evaluation}
\label{s:evaluation}

We seek to answer two key questions about our design; (1) how much benefit in accuracy does the AV receive from using \sysname and how is this accuracy impacted by the specific design choices we make, and (2) is our approach technically and economically feasible? 

We evaluate in two stages. First, in \cref{sec:eval-performance} we explore our system performance on a range of network conditions in simulation using thousands of real-world production AV traces from the Waymo Open Dataset~\cite{waymo-open-dataset} described in \cref{s:design:tasks}. As the scenarios are pre-recorded real-world data, we cannot modify the car's action during its drive, or detect a change in frequency of ``disengagements''~\cite{ca-dmv-disengagement-reports} or crashes~\cite{waymo-safety-impact,waymo-outperforms-humans} as reported in safety cards by real-world AV fleet operators. However, we are able to quantify exactly how much service accuracy increases on the real-world sensor data, which leads directly to improved environment perception and thus more faithful planning. We show in \cref{s:motivation} examples from the Waymo dataset where better models are able to detect a mid-distance pedestrian that the on-car model simply misses. Second, in \cref{s:eval:performance-real-world} we outfit a car with cameras, a mobile hotspot, a local server, and a cloud-hosted server, and run \sysname during a testdrive, reporting the system's real-world performance.

In all settings, we select the identity function $f_s(x)=x$ as our ``re-weighting function'' for all services (\cref{s:method:application-utility}) \ie maximizing average accuracy across services, however we note that alternative prioritization policies are possible.

\subsection{Simulation Performance}
\label{sec:eval-performance}

\begin{figure}
    \centering
    \includegraphics[width=0.6\linewidth]{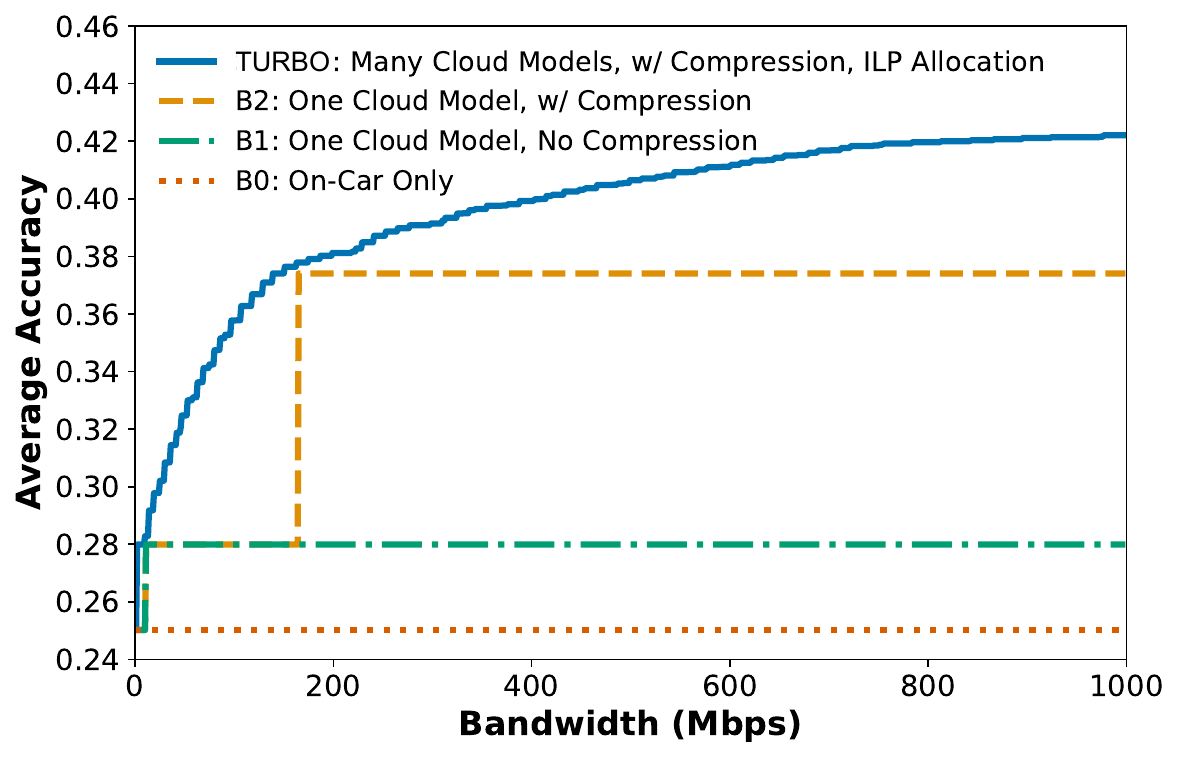}
   % \vspace{-0.5em}
    \caption{Average accuracy across services for \sysname compared to baselines of varying naivete as bandwidth increases. We assume an RTT of 20ms, object detection SLOs of 150ms, and motion planning SLO of 250ms, averaging performance across all scenarios.}
    \label{fig:utility-allocations}
    %\vspace{-1em}
\end{figure}

Our simulation control pipeline contains six services (as described in \cref{s:design:tasks}); five detection
services running for each of the five camera views available in the dataset,
as well as the motion prediction service.\footnote{
  Motion planning depends
on results of object detection; in this work, we assume pipelined execution
of the AV control
program~\cite{gog2021pylot,cruise-roscon,autoware-concepts}, thus bundle
motion planning using the results of earlier perception along with
perception on newly collected data. While it is conceivable to optimize
further by moving multiple sequential dependencies to the cloud to avoid
round-trips to the car, it is out of scope for our work; we leave such an
investigation to future work.}
\xspace We evaluate on-car and cloud
performance on the hardware reported in \cref{s:design:resources},
measuring the pre-processing, compression, decompression, and model runtime
for each model configuration on both the car and cloud hardware, then
constructing the end-to-end runtime of each selected configuration across
car and cloud.

\noindent\textbf{\sysname Accuracy vs. Baselines.} \cref{fig:utility-allocations} presents our method's performance compared to three tiers of baselines (B) of varying naivete as we vary the available network bandwidth, with a fixed RTT of 20ms (per~\cite{narayanan20215g-performance} measured for a server within 500km). B0 shows accuracy when running only on the car as is standard today, using the highest-accuracy model for each service that can run within the service SLO (ED1 for obj. detect per camera, a CNN-based model for motion planning). B1 shows the accuracy resulting from offering one cloud model for each service (ED4 for obj. detect per camera, a transformer-based model for motion planning) in addition to the on-car models running as backup, as proposed in prior work~\cite{schafhalter2023leveraging}. Finally, B2 applies object detection input preprocessing and compression (ED4 with preprocessing on-car, and compressed JPEG at 90\% quality) on the car before transferring over the network, still without special consideration to bandwidth allocation across services. 

\sysname (solid blue) improves on these baselines by up to 15.6\%pt by being bandwidth and utility-aware; this allows us to both allocate bandwidth optimally, and select the right preprocessing and compression to run for our model choice, based on the available bandwidth. We observe that in B1, after the initial bump due to the small-input-size motion planning service being able to move to the cloud, there is no benefit to any of the camera services until well over 2Gbps (not shown) as without special attention to bandwidth allocation as we propose, bandwidth is split equally amongst the services by TCP~\cite{chiu1989-fair-share}. B2 simply shifts this big step up earlier without fixing the underlying issue, and still provides only a single ``step'' point. In contrast, \sysname is able to quickly allocate bandwidth to the services that see the largest benefit (such as motion planning, corresponding to the jump at the left of the graph, as its inputs are relatively small and accuracy gains relatively big per \cref{s:design:motion-prediction}), then continually allocate bandwidth as necessary to achieve maximal accuracy.

\begin{figure*}[t]
    \centering
    \begin{minipage}[t]{0.48\textwidth}
        \centering
        \includegraphics[width=\linewidth]{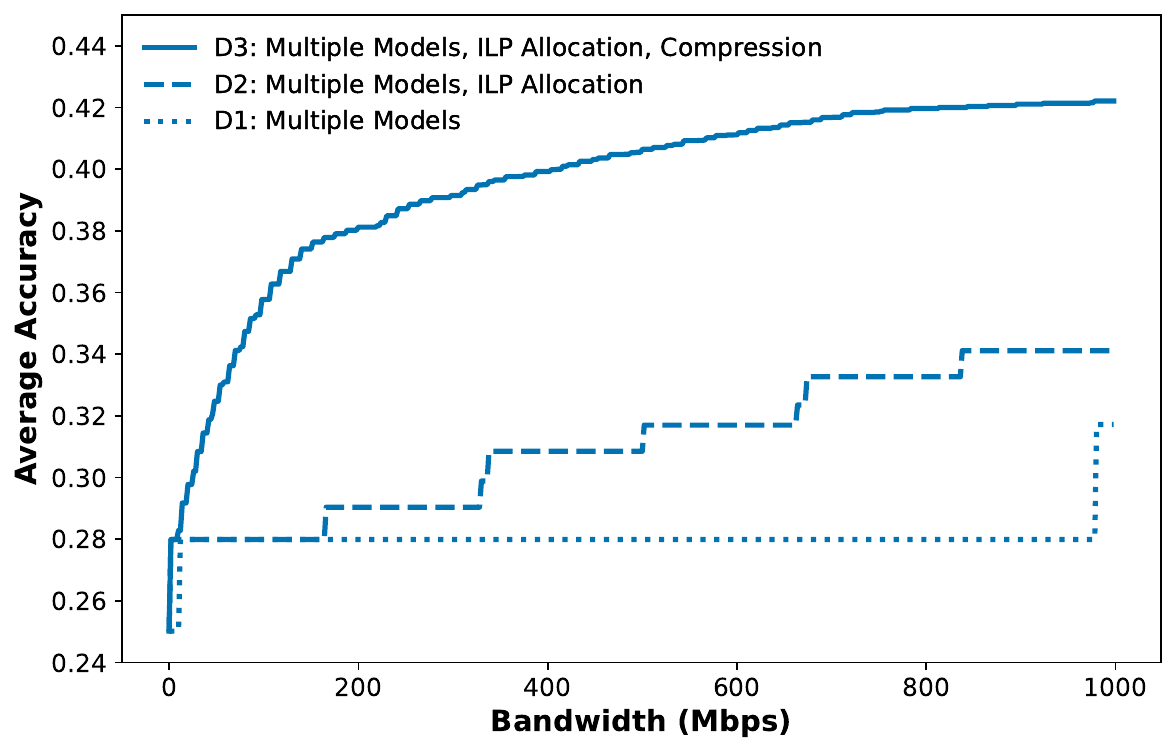}
        \caption{Incremental benefit from \sysname's design decisions.
        D1 naively uses multiple models to increase accuracy.
        D2 uses an ILP to effectively distribute bandwidth to services.
        D3 uses compression to reduce data transmission size.
        }
        \label{fig:alloc-methods-breakdown}
    \end{minipage}\hfill
    \begin{minipage}[t]{0.48\textwidth}
        \centering
        \includegraphics[width=\linewidth]{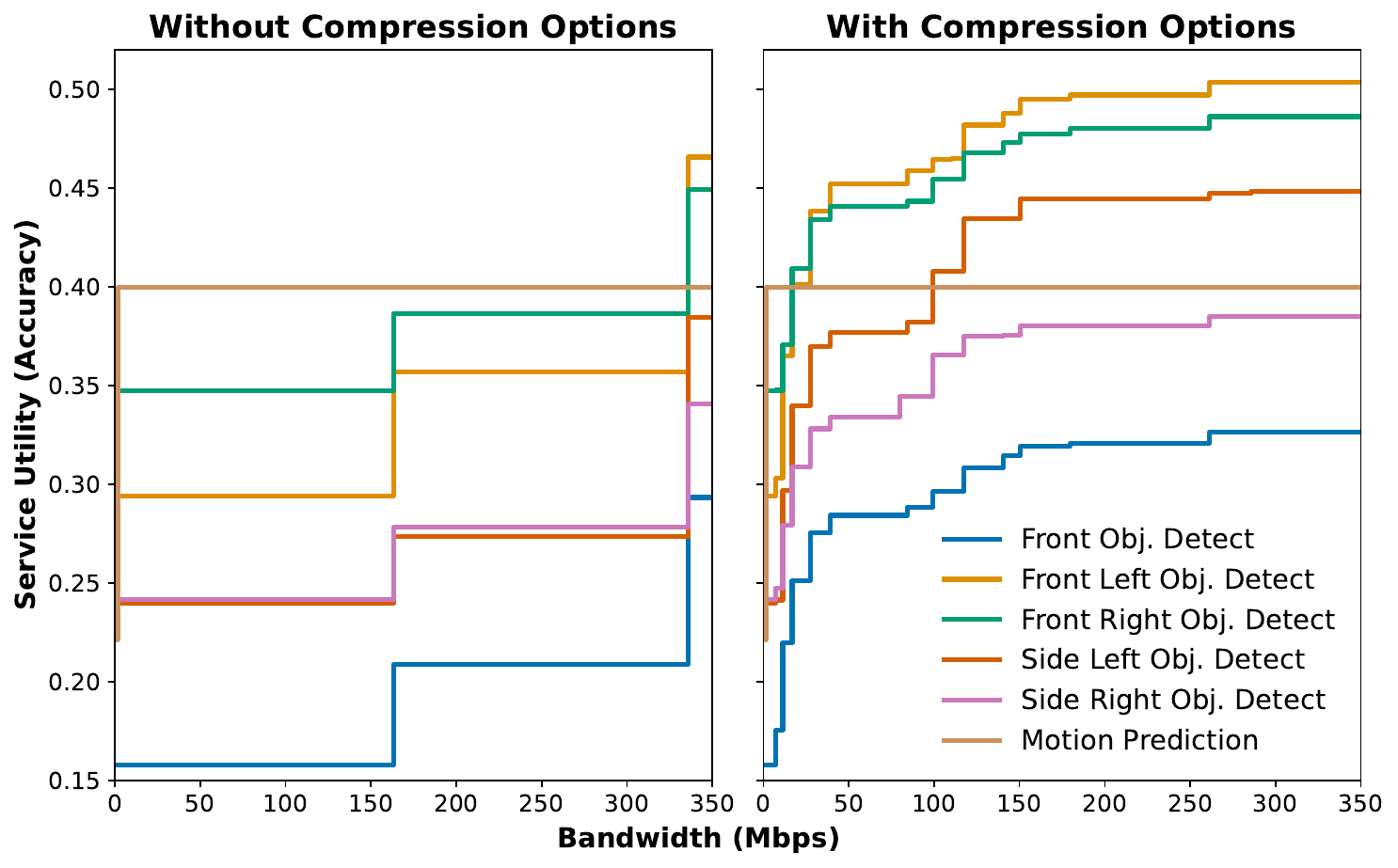}
        \caption{Utility curves for each service as derived by \sysname, without (left) and with (right) compression options considered, showing each service's accuracy if allocated a certain amount of bandwidth.}
        \label{fig:utility-curves-side-by-side}
    \end{minipage}
\end{figure*}

\noindent \textbf{Factor Analysis} \sysname makes three key design ideas to get its improved performance: (D1) adding multiple models to select from based on the available bandwidth, (D2) formulating that selection as a utility maximization ILP, and (D3) dynamically applying one of a range of different compression configurations. \cref{fig:alloc-methods-breakdown} shows the relative contribution of each.

\sysname's gains are attributed to the synthesis of the three; providing multiple cloud models of varying sizes per service allows services an earlier feasible step up to better models (980Mbps in \cref{fig:alloc-methods-breakdown}), however that happens in lock-step as bandwidth is shared equally across services by default. \sysname's ILP utility maximization formulation for bandwidth allocation allows individual services to "upgrade" cloud models in an order that it optimal for the AV's overall utility. Finally, adding compression provides many more opportunities at much lower bandwidths to step up to better cloud models, which provides the ILP formulation with many more points on which to optimize, as shown in \cref{fig:utility-curves-side-by-side}.

% We thus observe that the benefit of our system's management of bandwidth is two-fold; (1) by controller bandwidth allocation, we can apportion bandwidth to the services processing data that is "more important" and will benefit more from higher bandwidth, and (2) by being \textit{aware} of the bandwidth available to a service, we are able to choose the best model and compression scheme given available model runtimes and the network time their inputs would consume. Hence, we still see benefit from (2) even when we apply simpler bandwidth allocation algorithms. Further, our multiple compression options allow for more and earlier chances to step up to better models as bandwidth becomes available, as shown in \cref{fig:utility-curves-side-by-side}.

% \cref{fig:utility-curves-no-compression} and \cref{fig:utility-curves-with-compression} show the resultant derived utility curves for each service for the multi-model offerings without and with compression respectively.
\vspace{-0em}
\noindent\textbf{Varying Network Conditions} \sysname provides variable benefit based on the network conditions available. \cref{f:heatmap} shows the mean increase in accuracy across all scenarios for a range of bandwidth and RTT network conditions. While higher bandwidth and lower latency always provides more benefit, we see up to 10\%pt higher accuracy with as little as 150Mbps with RTTs of 20ms, well within the experienced operating ranges provided by 5G~\cite{5gdeploymentspecs,narayanan20215g-performance,ghoshal2022depth}.

\subsubsection{Dynamic Utility}
\begin{figure}[ht]
    \centering
    \begin{minipage}{0.48\linewidth}
        \centering
        \includegraphics[width=\linewidth]{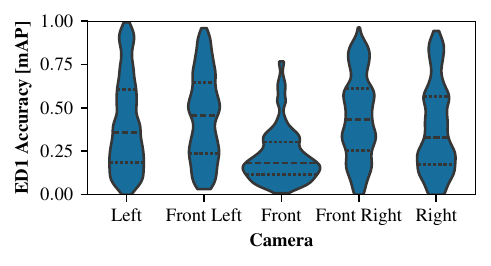}
        \caption{\cameraready{Distribution of accuracy across all frames for each camera service when using the ED1 model. The accuracy distributions vary greatly, as the distribution of object types and scene properties differ based on  positioning.}}
        \label{fig:accuracy-distribution}
    \end{minipage}\hfill
    \begin{minipage}{0.48\linewidth}
        \centering
        \includegraphics[width=\linewidth]{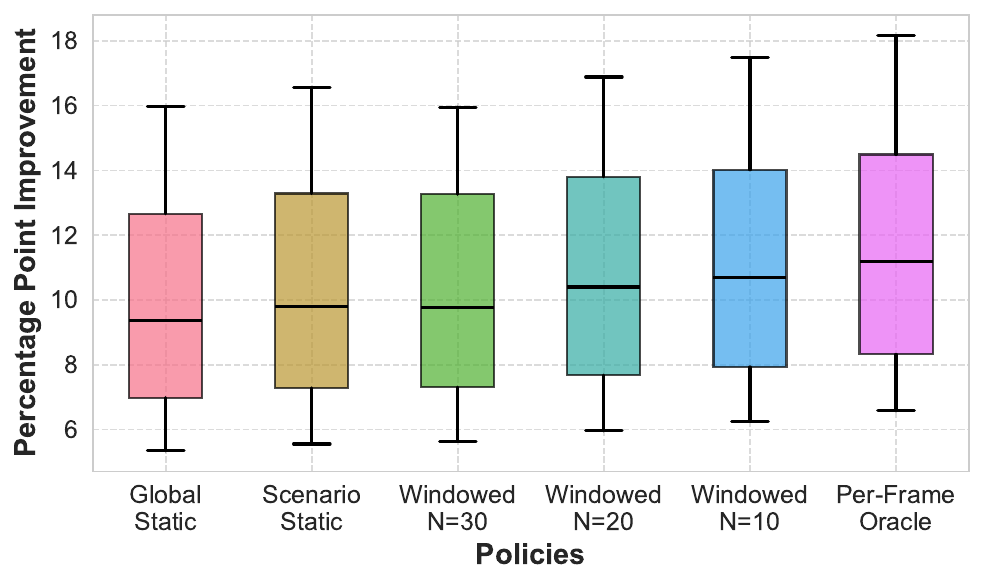}
        \caption{\cameraready{Distribution of \%pt improvement over on-car-only models for each policy, ranging from most static on the left to most dynamic on the right, assuming 250Mbps of bandwidth and 20ms RTT. 5, 25, 50, 75, 95th percentiles shown.}}
        \label{fig:true-mean-accuracies-by-policy}
    \end{minipage}
\end{figure}

\cameraready{%
In performing our analysis, we observed that a model's true accuracy can vary quite widely across frames, and further that these distributions vary greatly across services and camera angles (\cref{fig:accuracy-distribution}). To this end, we explored adapting our method using \textit{dynamic} utility curves rather than using the average accuracy calculated from an offline dataset. Our dynamic utility curves are derived from a recent group of recent frames in order to adapt to the current driving environment. We evaluate four different policies for generating dynamic utility curves with varying degrees of freshness:

\begin{enumerate}%[itemsep=-3pt] this broke formatting?
    \item \textbf{Global Static}: static utility curves derived from average accuracy across all frames of all scenarios, used above.
    \item \textbf{Scenario Static}: utility curves static to each scenario
      in the dataset, derived from the average accuracy  across all frames
      of the particular scenario.
    \item \textbf{N-Windowed}: utility curve computed from the
      $\text{N}^{\text{th}}$ frame of each scenario, used for the next
      $\text{N}-1$ frames. We experiment with $\text{N}=10,20,30,50$.
    \item \textbf{Per-Frame Oracle}: optimal ground-truth utility curve for each frame for each scenario, using the actual accuracy of each model on that frame.
\end{enumerate}

\cref{fig:true-mean-accuracies-by-policy} shows the distribution of accuracy improvement over using on-car only models. The Per-Frame Oracle policy serves as a performance upper-bound, showing the best-possible accuracies one could achieve with full information: an upper bound of +11.30\%pt median improvement over the on-vehicle models and +1.84\%pt over the global static policy.

However, recomputing utility curves for every frame is not practical.
A more practical dynamic policy such as the windowed N=20 policy, provides a 10.51\%pt % additional +1.05\%pt improvement at the median over static.
median improvement over the on-vehicle models and +1.05\%pt over global static.
Thus, we conclude that dynamically adjusting utility provides a small but meaningful increase in accuracy over our static policy, indicating that AVs can benefit from adjusting bandwidth allocations to account for changes in model accuracy when driving in different environments.}

\noindent \textbf{Accurate estimation of ground truth.} \cameraready{While the ``global
static'' curve can be computed on a large collection of previously collected
data, dynamic curves require some way of estimating model accuracy in a
given environment and window. Finding methods for doing so accurately is an
open problem~\cite{context-aware-streaming},
and thus beyond scope of the work.
%We do however observe that our requirement is weaker; simply determining relative accuracy difference between the models.
We observe that our requirement is weaker and that we can simply determine relative accuracy difference between the models.
To this end, we propose one way may be to periodically upload high-resolution images on spare bandwidth to run the best model and all models, and estimate relative accuracy by the magnitude of difference from a state-of-the-art model. Evaluating strategies for relative accuracy estimation is out of scope for this work. For our system implementation and real-world performance testing, we used the pre-computed global static policy.}

\subsection{Real-World Performance}
\label{s:eval:performance-real-world}

We run \sysname on an actual vehicle under real-world operating conditions, aiming to answer (1) what are \sysname's performance characteristics in the face of fluctuating network conditions and (2) how well actual accuracy benefit matches the system's expected benefit?

\begin{figure}
    \centering
    \begin{subfigure}{0.48\linewidth}
        \centering
        \includegraphics[width=\linewidth]{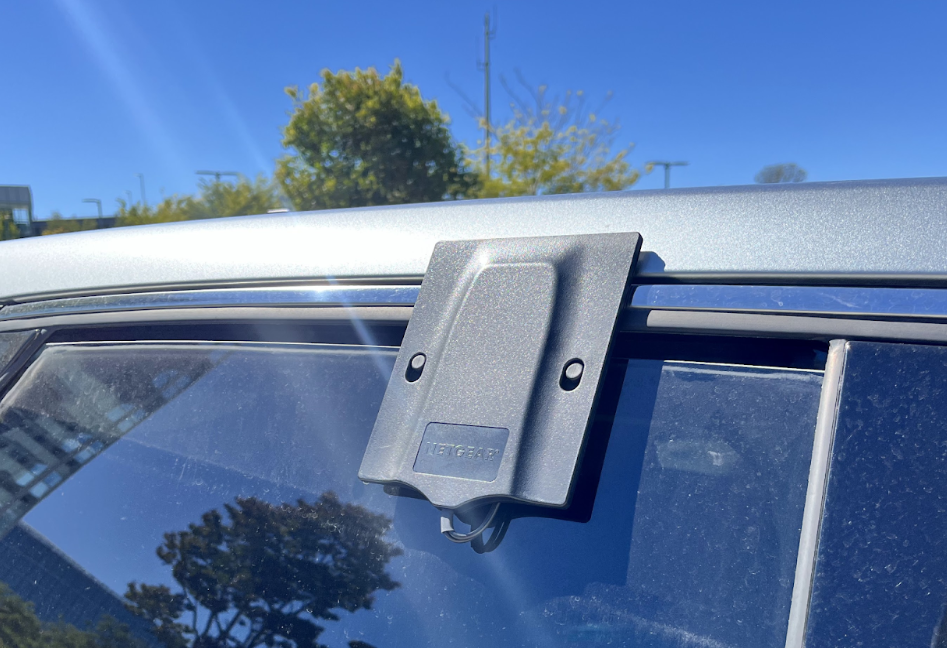}
        \caption{\cameraready{External view of the test car's window-mounted antenna.}}
        \label{fig:car-image-antenna}
    \end{subfigure}\hfill
    \begin{subfigure}{0.48\linewidth}
        \centering
        \includegraphics[width=\linewidth]{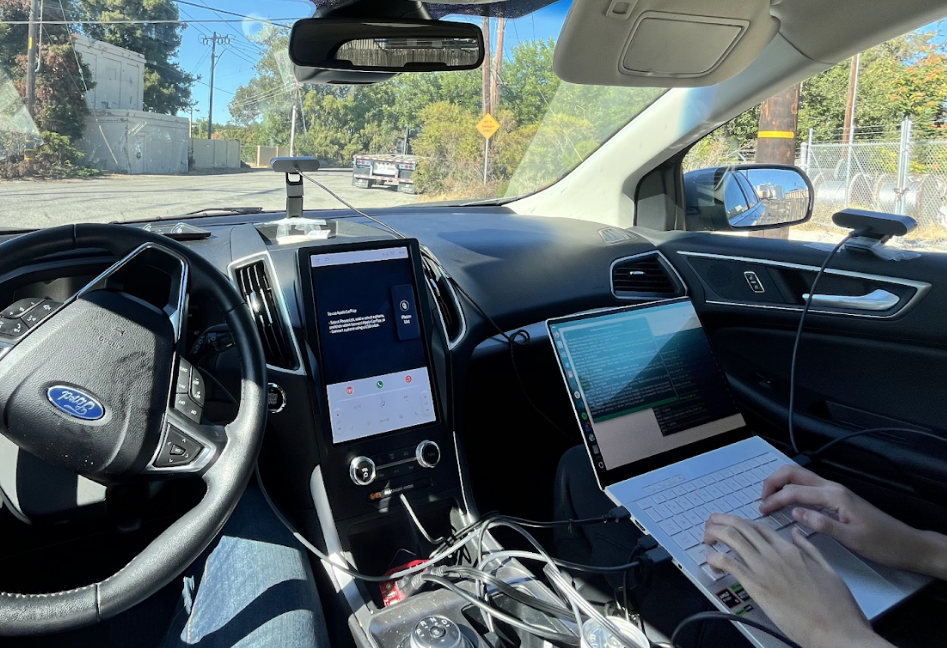}
        \caption{\cameraready{Interior view showing the mounted camera and local laptop server.}}
        \label{fig:car-image-inside}
    \end{subfigure}
    \caption{\cameraready{Experimental vehicle setup.}}
    \label{fig:car-setup}
\end{figure}

\noindent \textbf{Experiment Setup.} We outfitted our test car, a 2024 Ford Explorer, with the following hardware \cameraready{(shown in \cref{fig:car-setup})}; 3x Logitech Brio 4k webcams capturing front, right, and left videos streams; a consumer-grade NETGEAR NightHawk M6 Pro Mobile Hotspot with a T-Mobile SIM card; an omnidirectional window-mounted MIMO cellular antenna; an ROG Zephyrus G14 laptop with 8-core 4GHz processor and an NVIDIA GeForce RTX 4060 GPU, with direct ethernet connection to the hotspot; a 300W car plug power inverter to power the laptop and hotspot. For our remote server, we reserved an H100 GPU on Google Cloud; as nearby GPUs were unavailable at the time of our experiment, we reserved the closest available GPU 600 miles away.%, contributing to inflated RTT latencies.

We drove for 2 hours in a medium-sized metropolitan area in the United States, alternating between freeways and neighborhoods. Our car control system was responsible for ingesting and processing the data streams coming from the 3 cameras, running object detection models from the EfficientDet family we evaluated in the previous section, with the same 150ms SLO we use in simulation. Notably, we did not allow our control system to touch any car controls for safety reasons – the car was always fully driven by the authors, with the control system responsible for collecting and processing input as the “perception” control stage.

\begin{figure*}
    \centering
    \begin{subfigure}[b]{0.3\linewidth}
        \centering
        \includegraphics[width=\linewidth]{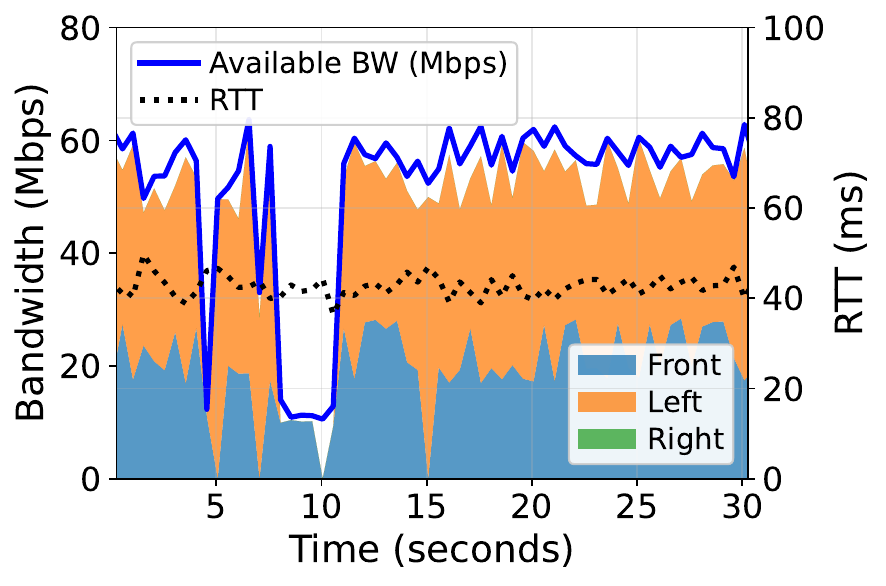}
        \caption{Network conditions and bandwidth allocations to each camera (service).}
        \label{fig:live-system-allocations}
    \end{subfigure}
    \hfill
    \begin{subfigure}[b]{0.3\linewidth}
        \centering
        \includegraphics[width=\linewidth]{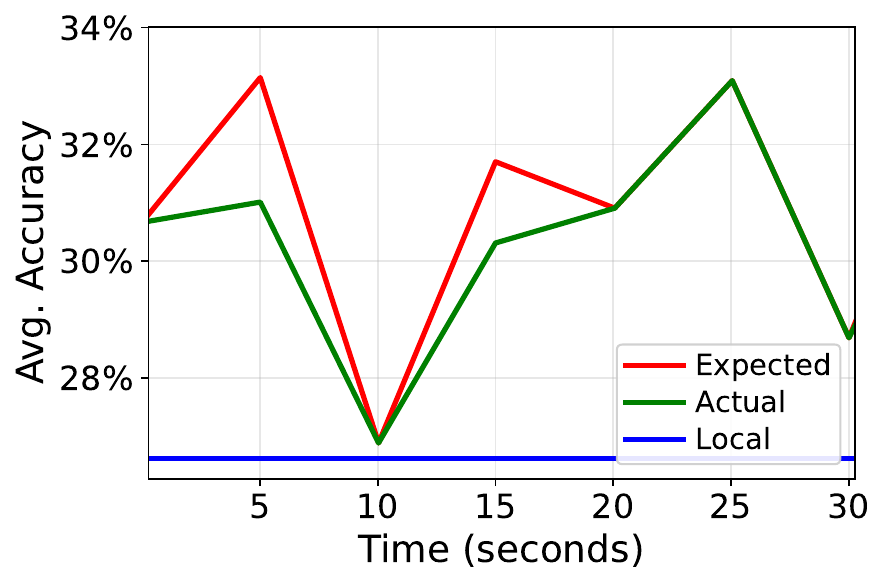}
        \caption{Average accuracy across services, both predicted and actually completed.}
        \label{fig:live-system-performance}
    \end{subfigure}
    \hfill
    \begin{subfigure}[b]{0.37\linewidth}
        \centering
        \includegraphics[width=\linewidth]{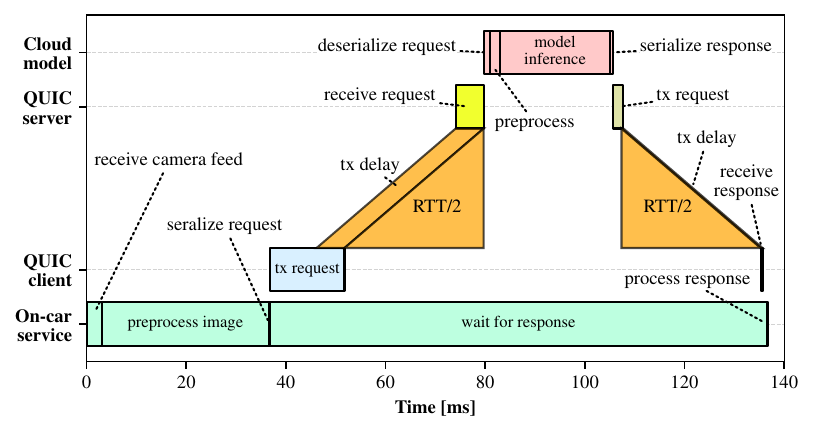}
        \caption{Trace of an object detector service where the
        propagation delay (RTT/2) dominates the response time.}
        \label{fig:live-system-benchmarks}
    \end{subfigure}
    %\vspace{-0.5em}
    \caption{\textbf{Production runs.} We deploy \sysname on a 5G-connected vehicle
    and measure its efficacy across three object detection services for the front, left, and right cameras.
    We find that \sysname effectively allocates the available bandwidth to its services
    (\cref{fig:live-system-allocations}) to increase the overall accuracy averaged across
    all services (\cref{fig:live-system-performance}).
    The trace of a single service (\cref{fig:live-system-benchmarks}) confirms that network
    latency is a key bottleneck and that our implementation adds minimal overhead.}
    \label{fig:system-benchmarks}
    %\vspace{-1em}
\end{figure*}

\noindent \textbf{Real-world utility from \sysname.} \cref{fig:live-system-allocations} shows a 30-second view of \sysname's allocations to each camera during a period of good network availability. During the wider 5-minute period surrounding this view, we saw 88\% of requests successfully returned, enabling the front camera to upgrade its model 76\% of the time, the left camera 86\%, and the right camera 0\%.%\footnote{We note this is the case under lower network conditions because the right camera sees comparatively lesser benefit from larger models than other cameras, as it sees fewer cars. This is because cars drive on the right in the US, so the left and front cameras see oncoming traffic, and the right does not. The camera's utility curve captures this accordingly.}
\footnote{\cameraready{We note that under constrained bandwidth conditions, our method prioritized upgrading the left and front cameras over the right because the right camera sees fewer cars (due to driving on the right), so benefits less from better models. This is captured implicitly in the utility curves we derived offline.}}
Across the three services, this translated to an absolute average increase of 4.1\% pointers of accuracy. We consume a total of 105.6 MB of data over this window. \cref{fig:live-system-performance} shows \sysname's predicted average accuracy, vs the actual accuracy when factoring in missed SLOs causing fallback to on-car models. 

Overall, we find our system successfully dynamically adapts to network conditions to make best use of network conditions, resulting in successful cloud offloads when network conditions allow. In times when cellular connectivity was poor, \sysname successfully maintained control pipeline integrity in the face of degraded network conditions, \textit{always} proceeding with the control pipeline within SLO using local results when network connectivity was unreliable. 

\noindent \textbf{System microbenchmarks.} We show a breakdown of the time spent in each component of our system (\cref{s:design:runtime-impl}) from the perspective of a single
service in \cref{fig:live-system-benchmarks}.
We find that our implementation provides minimal overhead and spends only 2.3 ms on serialization.
In this trace, the propagation delay (RTT) of 58.9 ms is the largest source of latency, which may be caused by congestion on the 5G network and the large distance to our server.
In contrast, the low transmission delay (5.6 ms upload, 0.4 ms download) is explained by
the small size of the compressed image (19 KB). 

\cameraready{We additionally profiled the runtime of our bandwidth allocation module, which we configured to run every 500ms as we updated our network measurements. We found the mean runtime of our optimization formulation running on the on-vehicle compute to be 14ms, with a standard deviation of 1ms and a maximum of 23ms. Thus, the computational overhead of solving the ILP is minimal, rather the bottleneck is the frequency of measuring network conditions.}

\noindent \textbf{Network performance.} In our run, we experienced worse network performance than expected: bandwidth conditions ranged from 0.80 Mbps (5th pct) to 55.86 Mbps (95th pct), and RTTs were high, starting at 60ms at the 5th pct and going up to hundreds of ms at the 95th pct. We suspect issues with the antenna for our setup, as literature reports considerably higher network performance: 5G deployments target a \textit{minimum (5th-percentile)} user-experienced bandwidth of 50 Mbps for uplink and 100 Mbps for downlink~\cite{5gdeploymentspecs}, and prior work shows moving vehicle 5G network conditions to be expected at 80 Mbit/s (25th pct) to 160 Mbit/s (75th pct) for uplink bandwidth, and 12ms (25th pct) to 18ms (75th pct) for base station RTT ping latency~\cite{ghoshal, razzaghpour}.

%and 200 Mbit/s (25th pct) to 250 Mbit/s (75th pct) for downlink

\noindent \textbf{Cost.}
For our test-drive, we purchased an unlimited plan from a carrier with a monthly high-speed usage cap at 50GB. Our test run did not exceed this cap, however a continuous real-world deployment would need to purchase more data to run continuously throughout the month. In \cref{s:appendix-feasibility}, we estimate our total hourly cost of remote resources at \$5.27, with \$2.78 from the network (at the 10th global percentile) and \$2.49 from compute for an H100. We emphasize that the true cost of cloud compute is likely lower due to better efficiency when operating at scale.
Our method is tunable for cost-sensitive markets: operators can tune their utility curves to take into account cost-benefit, \eg by executing on cheaper GPUs and selectively using cloud models in challenging environments.
\section{Discussion}
\label{s:discussion}

\textbf{Limitations.} Our design does not consider scheduling the order of messages,
\ie delaying transmission for one service to make more bandwidth available for another service.
% \ie delaying transmission for one service to make more bandwidth available to another service to reduce its transmission time.
Likewise, we do not consider executing consecutive tasks in the cloud without returning back to the car which could further reduce data transmission.
We focus on offloading parallelizable tasks, and leave scheduling extensions (\eg exploiting task dependencies) to future work. \sysname relies on 5G coverage for fast network bandwidths and low round-trip times. While 5G coverage is expanding,
it is not available in every region~\cite{ericsson-5gcoverage}.
Finally, \sysname requires the AV operator to configure $f_s$ (\cref{s:method:application-utility}) to ensure that the ILP maximizes the end-to-end performance of the AV system, as maximizing average accuracy across \textit{all} services may result in maximal safety.
We believe that configurations of $f_s$ may be discovered using simulations and machine learning, and leave this area to future work.

% \myparagraph{Additional points to cover (possibly elsewhere).}
% \begin{itemize}
    % \item Bandwidth allocation happens on-vehicle.
    % \item Future work: bandwidth allocation for multi-car setting.
    % \item Download transmission time. Say that the results are small, so this is low. Add as footnote.
% \end{itemize}

% \myparagraph{Limitations.}
% \begin{itemize}
%     \item Don't consider ordering of messages (or scheduling across time). We only schedule in the aggregate. Benefit: this could improve latency further.
%     \item Require 5G coverage for maximal benefit.
%     \item Requires operator to configure $f_s$ to prioritize services. Could be discovered in simulation using ML!
%     \item Cost estimate is based on consumer pricing (realism).
% \end{itemize}

% \subsection{Implications for Autonomous Vehicles}
% \begin{itemize}
%   \item Potential to increase safety via access to more accurate models.
%   \item Reduce remote interventions by running better models which reduces
%     vehicle confusion. This improves the experience for riders and reduces
%     the cost of operating AVs.
%   \item The cellular connecton becomes a key part of the AV's compute
%     infrastructure. This is a departure from the current model where the
%     AV is a self-contained unit, and may require better networks with
%     performance guarantees and security.
% \end{itemize}

\myparagraph{Implications for AV design.}
Our results suggest that using cloud resources to run more accurate models
can significantly improve accuracy, enabling AVs to make better-informed,
high quality decisions which benefit safety.
Beyond safety events, cloud models can reduce the frequency of remote interventions~\cite{waymo-fleet-response} by improving the AV's understanding of its surroundings.
Fewer interventions improves the experience for riders, reduces the cost of operating AVs, and is an important metric tracked by government officials in assessing AV system safety~\cite{ca-dmv-disengagement-reports}.

At the same time, we note that adopting the cloud raises new concerns; the cellular connection is elevated to a more important role, which increases the need for security in the AV system as well as
improvements to the performance, reliability, and availability of the cellular connection.

% Finally, adopting the cloud to run more accurate AV models promotes the
% cellular connection to a key part of the AV system's design.
%
% Finally, integrating the cloud into AV systems is a departure from their current design as self-contained units and elevates the importance of the celullar connection.
%
% This shift increases the need for security in the AV system as well as
% improvements to the performance, reliability, and availability of the
% cellular connection.

% \subsection{Implications for Mobile Network Stakeholders}
% \begin{itemize}
%   \item If cloud computing becomes a key component of autonomous driving,
%     cellular network demand will increase. There were 283 million vehicles
%     registered in the US in 2022 and an estimated 1.5 billion vehicles
%     worldwide. Even if only a fraction of these vehicles are autonomous and
%     use the cloud, the added demand for network resources will be
%     significant.
%   \item More demand for network performance, in particular uplink bandwidth
%     and low round-trip latency. As a key bottleneck is the hop to the
%     cell tower (30-40 ms which makes up most of the RTT), edge servers
%     provide insufficient benefit.
%   \item Need to support, prioritize, and potentially reserve capacity for
%     safety-critical applications such as AVs. 
% \end{itemize}
\myparagraph{Implications for Mobile Network Stakeholders.}
If AVs integrate cloud computing, the demand for cellular network resources
will increase massively.
In 2022, 283 million vehicles were registered in the
United States~\cite{usa-vehicle-registrations-2022} alone out of an
estimated 1.5 billion vehicles in use
worldwide~\cite{bernstein-ev-revolution-2021}.
If only a fraction of these vehicles use the cloud to enhance autonomous
capabilities, mobile networks networks must support millions of new users
with long-running, data-intensive streaming workloads.
Moreover, AVs demand better uplink bandwidth and lower round-trip
latencies.
A key bottleneck to RTT is the hop from the 5G base station to the cellular
core network, which accounts for most of the RTT according to
\cite{xu2020understanding}.
Consequently, cellular network operators will need to prioritize
performance improvements and potentially provide network guarantees for AVs
and other safety-critical connected systems.

% \subsection{Applications Beyond Autonomous Driving}
% \begin{itemize}
%   \item Vehicles with limited autonomy. For example, smart parking garages,
%     slow-speed driving, beter driver assistance.
%   \item Drones. Drones are similar to AVs in that they are mobile devices
%     that require real-time processing of sensor data. They are even more
%     constrained in terms of power and compute resources because they need
%     to remain lightweight.
%   \item Robotics. Real-time processing of sensor data for robots in
%     warehouses, factories, and other settings.
%   \item Vision-Language models. Real-time processing of video feeds to
%     provide audio descriptions for visually impaired persons.
%   \item Security systems. Real-time processing of video feeds to detect
%     intruders or other security threats.
% \end{itemize}
\myparagraph{Applications beyond AVs.}
We believe that the benefits of using cloud hardware in real-time to boost
accuracy has far-reaching applications.
Beyond AVs, \sysname can benefit vehicles with limited autonomy, such as
hands-free highway driving or self-parking.
Similarly, autonomous drones use ML to process sensor data in real
time.
Drones exhibit even more stringent constraints on compute due to weight
restrictions, making access to cloud resources an attractive option.
Robotics~\cite{fogros}, security systems, and vision-language models are
other applications which benefit from real-time access to cloud resources
to benefit accuracy.
% \alex{additional future work: cross-car Bandwidth allocation at the tower level}\alex{should we state much more strongly here –– our approach is further applicable to any real-time edge control system?}

\section{Related Works}
\label{s:related-works}

\myparagraph{Autonomous driving with remote resources.}
Several works propose designs for decentralized systems which build on
Vehicle-to-Vehicle (V2V) or Vehicle-to-Infrastructure (V2X) communication
patterns to harness additional compute
resources~\cite{liu2019edge-assisted-detection,cui2020offloading,sun2018learning,sun2018cooperative},
share data~\cite{kumar2012cloud,zhang2021emp,qiu2018avr}, or develop
collaborative
algorithms~\cite{chen2019f,wang2017cooperative,ngo2023cooperative,liu2019trajectory}.
While such approaches can improve the accuracy of AV services, prior works
do not study accuracy-aware bandwidth allocation decisions
under fluctuating network conditions. \cameraready{One prior work~\cite{schafhalter2023leveraging} shows that runtimes of AV-specific models in the cloud are sufficiently faster than on typical on-car compute to make room for cellular ping latencies, enabling net quicker car reaction times. However, this work does not account for bandwidth-induced delay, only measuring \textit{ping} RTTs for the network delay. We show in \cref{sec:eval-performance} that naively including bandwidth-induced delay negates any cloud benefit.}

\myparagraph{Edge-cloud inference.}
How to partition a single neural network between mobile devices and
datacenters is an active area of
research~\cite{neurosurgeon,ra2011odessa,zhang2020towards}.
While existing approaches reduce inference latency and energy consumption,
they do not apply to real-time settings with multiple services and multiple
potential model configurations.
%
% Existing approaches partition a single neural network between mobile
% devices and datacenters~\cite{neurosurgeon,ra2011odessa,zhang2020towards} to reduce
% inference latency and energy consumption, but do not consider settings with
% multiple services and multiple model configurations.
In the cloud, model serving systems~\cite{clipper,infaas,clockwork} are
designed to serve many different model configurations while meeting
statistical SLOs, which is complementary to our work.
Because these systems are deployed in datacenters, they do not consider
bandwidth a key resource under contention which needs to be managed.
% \cameraready{One prior work~\cite{wang-infocomm2020-jointconfig} examines a similar formulation for joint inference configuration and bandwidth allocation, ...<PETER TO ADD TEXT HERE>.}

\myparagraph{Reconfigurable video analytics systems.}
Several systems reconfigure the ML inference pipeline to react to changes
in the input
content~\cite{videostorm,xu2022litereconfig,xu2022smartadapt,chameleon,liu2019edge}
and to maximize real-time accuracy
metrics~\cite{streaming-perception,context-aware-streaming}.
Video analytics systems like Reducto~\cite{li2020reducto} and
DDS~\cite{du2020server} maintain high accuracy while reducing edge-cloud
network traffic, but do not address % how to manage
contention for bandwidth.
Ekya~\cite{ekya} and NoScope~\cite{kang2017noscope} use online
learning to further improve the accuracy of video analytics.
%
% Peter: I think this fits better here, since JCAB cites some of these papers.
\cameraready{JCAB~\cite{wang-infocomm2020-jointconfig} proposes a similar joint optimization problem
to allocate bandwidth across multiple video streams and maximize overall accuracy.
In contrast to \sysname, JCAB targets real-time analytics instead of real-time safety-critical control.
As such, JCAB targets a long-term average latency SLO, does not factor in RTT,
and varies frame rates to trade-off accuracy and bandwidth instead of compression.
}

% \myparagraph{Video compression and streaming.}

\myparagraph{Bandwidth allocation algorithms.} The problem of utility-aware bandwidth allocation has been studied in a variety of concrete settings~\cite{nagaraj2016-numfabric,liu2007bandwidthalloc,huang2018sdnalloc}, as well as more abstractly~\cite{kelly1998ratecontrol,zegura1999utility}. Google's BwE paper~\cite{google2015bwe} takes a similar approach to ours of decomposing overall utility into per-service utility curves, however they compose and optimize over their per-service curves differently as their objective is max-min fairness across flows. Furthermore, they perform their allocation centrally across a global WAN with different tasks and users. We examine the problem of bandwidth allocation from the perspective of the components components within the AV control application, which is part of the larger class of compound AI systems~\cite{compound-ai-systems-blog}.

\newpage
\bibliography{bibs}

@article{claypool2017self,
  title={{Self-Driving Cars: The Impact on People with Disabilities}},
  author={Claypool, Henry and Bin-Nun, Amitai and Gerlach, Jeffrey},
  journal={Newton, MA: Ruderman Family Foundation},
  year={2017}
}

@article{katrakazas,
  author={Katrakazas, Christos and Quddus, Mohammed and Chen, Wen-Hua and Deka, Lipika},
  title={{Real-time Motion Planning Methods for Autonomous On-Road Driving: State-of-the-art and Future Research Directions}},
  journal={Transportation Research Part C: Emerging Technologies},
  volume={60},
  pages={416--442},
  year={2015},
}

@article{paden-survey,
  title={{A Survey of Motion Planning and Control Techniques for Self-Driving Urban Vehicles}},
  author={Paden, Brian and {\v{C}}{\'a}p, Michal and Yong, Sze Zheng and Yershov, Dmitry and Frazzoli, Emilio},
  journal={IEEE Transactions on Intelligent Vehicles},
  volume={1},
  number={1},
  pages={33--55},
  year={2016},
  publisher={IEEE}
}

@article{guanetti_control_2018,
  title = {Control of Connected and Automated Vehicles: {{State}} of the Art and Future Challenges},
  shorttitle = {Control of Connected and Automated Vehicles},
  author = {Guanetti, Jacopo and Kim, Yeojun and Borrelli, Francesco},
  year = {2018},
  month = jan,
  volume = {45},
  pages = {18--40},
  issn = {1367-5788},
  doi = {10.1016/j.arcontrol.2018.04.011},
  journal = {Annual Reviews in Control},
  language = {en}
}

@article{yolov4,
  author = {Bochkovskiy, Alexey and Wang, Chien-Yao and Mark Liao, Hong-Yuan},
  title = {{YOLOv4}: Optimal Speed and Accuracy of Object Detection},
  year = {2020},
  eprint = {arXiv:2004.10934},
  url = {https://arxiv.org/abs/2004.10934},
}

@article{waymo-open-dataset,
  author = {Pei Sun and Henrik Kretzschmar and Xerxes Dotiwalla and Aurelien Chouard and Vijaysai Patnaik and Paul Tsui and James Guo and Yin Zhou and Yuning Chai and Benjamin Caine and Vijay Vasudevan and Wei Han and Jiquan Ngiam and Hang Zhao and Aleksei Timofeev and Scott Ettinger and Maxim Krivokon and Amy Gao and Aditya Joshi and Sheng Zhao and Shuyang Cheng and Yu Zhang and Jonathon Shlens and Zhifeng Chen and Dragomir Anguelov},
  title = {Scalability in Perception for Autonomous Driving: Waymo Open Dataset},
  year = {2019},
  eprint = {arXiv:1912.04838},
  url = {https://arxiv.org/abs/1912.04838}
}

@article{neurosurgeon,
  title={Neurosurgeon: Collaborative intelligence between the cloud and mobile edge},
  author={Kang, Yiping and Hauswald, Johann and Gao, Cao and Rovinski, Austin and Mudge, Trevor and Mars, Jason and Tang, Lingjia},
  journal={ACM SIGARCH Computer Architecture News},
  volume={45},
  number={1},
  pages={615--629},
  year={2017},
  publisher={ACM New York, NY, USA}
}

@article{vit,
  title={An image is worth 16x16 words: Transformers for image recognition at scale},
  author={Dosovitskiy, Alexey and Beyer, Lucas and Kolesnikov, Alexander and Weissenborn, Dirk and Zhai, Xiaohua and Unterthiner, Thomas and Dehghani, Mostafa and Minderer, Matthias and Heigold, Georg and Gelly, Sylvain and others},
  journal={arXiv preprint arXiv:2010.11929},
  year={2020}
}

@article{wolfe2020rapid,
  title={Rapid holistic perception and evasion of road hazards.},
  author={Wolfe, Benjamin and Seppelt, Bobbie and Mehler, Bruce and Reimer, Bryan and Rosenholtz, Ruth},
  journal={Journal of experimental psychology: general},
  volume={149},
  number={3},
  pages={490},
  year={2020},
  publisher={American Psychological Association}
}

@article{johansson1971drivers,
  title={Drivers' brake reaction times},
  author={Johansson, Gunnar and Rumar, K{\aa}re},
  journal={Human factors},
  volume={13},
  number={1},
  pages={23--27},
  year={1971},
  publisher={SAGE Publications Sage CA: Los Angeles, CA}
}

@article{liu2019trajectory,
  title={Trajectory planning for autonomous intersection management of connected vehicles},
  author={Liu, Bing and Shi, Qing and Song, Zhuoyue and El Kamel, Abdelkader},
  journal={Simulation Modelling Practice and Theory},
  volume={90},
  pages={16--30},
  year={2019},
  publisher={Elsevier}
}

@article{chauffeurnet,
  title={Chauffeurnet: Learning to drive by imitating the best and synthesizing the worst},
  author={Bansal, Mayank and Krizhevsky, Alex and Ogale, Abhijit},
  journal={arXiv preprint arXiv:1812.03079},
  year={2018}
}

@article{wu2023point,
  title={Point transformer v3: Simpler, faster, stronger},
  author={Wu, Xiaoyang and Jiang, Li and Wang, Peng-Shuai and Liu, Zhijian and Liu, Xihui and Qiao, Yu and Ouyang, Wanli and He, Tong and Zhao, Hengshuang},
  journal={arXiv preprint arXiv:2312.10035},
  year={2023}
}

@article{shi2024mtr++,
  title={MTR++: Multi-agent motion prediction with symmetric scene modeling and guided intention querying},
  author={Shi, Shaoshuai and Jiang, Li and Dai, Dengxin and Schiele, Bernt},
  journal={IEEE Transactions on Pattern Analysis and Machine Intelligence},
  year={2024},
  publisher={IEEE}
}

@misc{leng2024pvtransformer,
  title={PVTransformer: Point-to-Voxel Transformer for Scalable 3D Object Detection}, 
  author={Zhaoqi Leng and Pei Sun and Tong He and Dragomir Anguelov and Mingxing Tan},
  year={2024},
  eprint={2405.02811},
  archivePrefix={arXiv},
  primaryClass={cs.CV}
}

@misc{mu2024most,
  title={MoST: Multi-modality Scene Tokenization for Motion Prediction}, 
  author={Norman Mu and Jingwei Ji and Zhenpei Yang and Nate Harada and Haotian Tang and Kan Chen and Charles R. Qi and Runzhou Ge and Kratarth Goel and Zoey Yang and Scott Ettinger and Rami Al-Rfou and Dragomir Anguelov and Yin Zhou},
  year={2024},
  eprint={2404.19531},
  archivePrefix={arXiv},
  primaryClass={cs.CV}
}

@article{honda2023recall, 
  url={https://hondanews.com/en-US/honda-corporate/releases/release-13f25e90cfe47cd58453f1f710208486-multi-model-brake-master-cylinder-recall},
  title = {Multi-model Brake Master Cylinder Recall},
  author = {Honda},
  journal={Honda News},
  publisher={Honda},
  year={2023},
  month={Jul}
}

@inproceedings{yin2021center,
  title={Center-based 3d object detection and tracking},
  author={Yin, Tianwei and Zhou, Xingyi and Krahenbuhl, Philipp},
  booktitle={Proceedings of the IEEE/CVF conference on computer vision and pattern recognition},
  pages={11784--11793},
  year={2021}
}

@article{tang-lane-detection-review,
  title = {A review of lane detection methods based on deep learning},
  journal = {Pattern Recognition},
  volume = {111},
  pages = {107623},
  year = {2021},
  issn = {0031-3203},
  doi = {https://doi.org/10.1016/j.patcog.2020.107623},
  url = {https://www.sciencedirect.com/science/article/pii/S003132032030426X},
  author = {Jigang Tang and Songbin Li and Peng Liu},
}

@article{shi2022motion,
  title={Motion transformer with global intention localization and local movement refinement},
  author={Shi, Shaoshuai and Jiang, Li and Dai, Dengxin and Schiele, Bernt},
  journal={Advances in Neural Information Processing Systems},
  volume={35},
  pages={6531--6543},
  year={2022}
}

@misc{ettinger2024scaling,
      title={Scaling Motion Forecasting Models with Ensemble Distillation}, 
      author={Scott Ettinger and Kratarth Goel and Avikalp Srivastava and Rami Al-Rfou},
      year={2024},
      eprint={2404.03843},
      archivePrefix={arXiv},
      primaryClass={cs.RO}
}

@inproceedings{hu2023planning,
  title={Planning-oriented autonomous driving},
  author={Hu, Yihan and Yang, Jiazhi and Chen, Li and Li, Keyu and Sima, Chonghao and Zhu, Xizhou and Chai, Siqi and Du, Senyao and Lin, Tianwei and Wang, Wenhai and others},
  booktitle={Proceedings of the IEEE/CVF Conference on Computer Vision and Pattern Recognition},
  pages={17853--17862},
  year={2023}
}

@article{teng2023planning-survey,
  title={Motion planning for autonomous driving: The state of the art and future perspectives},
  author={Teng, Siyu and Hu, Xuemin and Deng, Peng and Li, Bai and Li, Yuchen and Ai, Yunfeng and Yang, Dongsheng and Li, Lingxi and Xuanyuan, Zhe and Zhu, Fenghua and others},
  journal={IEEE Transactions on Intelligent Vehicles},
  year={2023},
  publisher={IEEE}
}

@inproceedings{lu2023imitation,
  title={Imitation is not enough: Robustifying imitation with reinforcement learning for challenging driving scenarios},
  author={Lu, Yiren and Fu, Justin and Tucker, George and Pan, Xinlei and Bronstein, Eli and Roelofs, Rebecca and Sapp, Benjamin and White, Brandyn and Faust, Aleksandra and Whiteson, Shimon and others},
  booktitle={2023 IEEE/RSJ International Conference on Intelligent Robots and Systems (IROS)},
  pages={7553--7560},
  year={2023},
  organization={IEEE}
}

@article{ngo2023cooperative,
  title={Cooperative perception with V2V communication for autonomous vehicles},
  author={Ngo, Hieu and Fang, Hua and Wang, Honggang},
  journal={IEEE Transactions on Vehicular Technology},
  volume={72},
  number={9},
  pages={11122--11131},
  year={2023},
  publisher={IEEE}
}

@article{liu2019edge,
  title={Edge computing for autonomous driving: Opportunities and challenges},
  author={Liu, Shaoshan and Liu, Liangkai and Tang, Jie and Yu, Bo and Wang, Yifan and Shi, Weisong},
  journal={Proceedings of the IEEE},
  volume={107},
  number={8},
  pages={1697--1716},
  year={2019},
  publisher={IEEE}
}

@article{konev2022motioncnn,
  title={Motioncnn: A strong baseline for motion prediction in autonomous driving},
  author={Konev, Stepan and Brodt, Kirill and Sanakoyeu, Artsiom},
  journal={arXiv preprint arXiv:2206.02163},
  year={2022}
}

@article{chiu1989-fair-share,
title = {Analysis of the increase and decrease algorithms for congestion avoidance in computer networks},
journal = {Computer Networks and ISDN Systems},
volume = {17},
number = {1},
pages = {1-14},
year = {1989},
issn = {0169-7552},
doi = {https://doi.org/10.1016/0169-7552(89)90019-6},
url = {https://www.sciencedirect.com/science/article/pii/0169755289900196},
author = {Dah-Ming Chiu and Raj Jain}
}

@article{yolov6,
  title={YOLOv6: A single-stage object detection framework for industrial applications},
  author={Li, Chuyi and Li, Lulu and Jiang, Hongliang and Weng, Kaiheng and Geng, Yifei and Li, Liang and Ke, Zaidan and Li, Qingyuan and Cheng, Meng and Nie, Weiqiang and others},
  journal={arXiv preprint arXiv:2209.02976},
  year={2022}
}

@inproceedings{wayformer,
  title={Wayformer: Motion forecasting via simple \& efficient attention networks},
  author={Nayakanti, Nigamaa and Al-Rfou, Rami and Zhou, Aurick and Goel, Kratarth and Refaat, Khaled S and Sapp, Benjamin},
  booktitle={2023 IEEE International Conference on Robotics and Automation (ICRA)},
  pages={2980--2987},
  year={2023},
  organization={IEEE}
}

@ARTICLE{nytimes2024selfdrivinghelp,
  title     = "How Self-Driving Cars Get Help From Humans Hundreds of Miles Away",
  author    = "Metz, Cade and Henry, Jason and Laffin, Ben and Lieberman,
               Rebecca and Lu, Yiwen",
  journal   = "The New York Times",
  publisher = "The New York Times",
  month     =  sep,
  year      =  2024,
  language  = "en"
}

@article{wang2006utilities,
author = {Wang, Wei-Hua and Palaniswami, Marimuthu and Low, Steven H.},
title = {Application-oriented flow control: fundamentals, algorithms and fairness},
year = {2006},
issue_date = {December 2006},
publisher = {IEEE Press},
volume = {14},
number = {6},
issn = {1063-6692},
url = {https://doi.org/10.1109/TNET.2006.886318},
doi = {10.1109/TNET.2006.886318},
journal = {IEEE/ACM Trans. Netw.},
month = {dec},
pages = {1282–1291},
numpages = {10}
}

@article{zhang2020towards,
  title={Towards real-time cooperative deep inference over the cloud and edge end devices},
  author={Zhang, Shigeng and Li, Yinggang and Liu, Xuan and Guo, Song and Wang, Weiping and Wang, Jianxin and Ding, Bo and Wu, Di},
  journal={Proceedings of the ACM on Interactive, Mobile, Wearable and Ubiquitous Technologies},
  volume={4},
  number={2},
  pages={1--24},
  year={2020},
  publisher={ACM New York, NY, USA}
}

@INPROCEEDINGS{liu2007bandwidthalloc,
  author={Liu, Changbin and Shi, Lei and Liu, Bin},
  booktitle={Fourth European Conference on Universal Multiservice Networks (ECUMN'07)}, 
  title={Utility-Based Bandwidth Allocation for Triple-Play Services}, 
  year={2007},
  volume={},
  number={},
  pages={327-336},
  doi={10.1109/ECUMN.2007.58}
}

@ARTICLE{huang2018sdnalloc,
  author={Huang, Xiaohong and Yuan, Tingting and Ma, Maode},
  journal={IEEE Access}, 
  title={Utility-Optimized Flow-Level Bandwidth Allocation in Hybrid SDNs}, 
  year={2018},
  volume={6},
  number={},
  pages={20279-20290},
  doi={10.1109/ACCESS.2018.2820682}
}

@INPROCEEDINGS{wang-infocomm2020-jointconfig,
  author={Wang, Can and Zhang, Sheng and Chen, Yu and Qian, Zhuzhong and Wu, Jie and Xiao, Mingjun},
  booktitle={IEEE INFOCOM 2020 - IEEE Conference on Computer Communications}, 
  title={Joint Configuration Adaptation and Bandwidth Allocation for Edge-based Real-time Video Analytics}, 
  year={2020},
  volume={},
  number={},
  pages={257-266},
  doi={10.1109/INFOCOM41043.2020.9155524}}

@inproceedings{erdos,
  title={D3: {A} {D}ynamic {D}eadline-{D}riven approach for {B}uilding {A}utonomous {V}ehicles},
  author={Gog, Ionel and Kalra, Sukrit and Schafhalter, Peter and Gonzalez, Joseph E and Stoica, Ion},
  booktitle={Proceedings of the Seventeenth European Conference on Computer Systems},
  pages={453--471},
  year={2022}
}

@inproceedings{context-aware-streaming,
  title={Context-Aware Streaming Perception in Dynamic Environments},
  author={Sela, Gur-Eyal and Gog, Ionel and Wong, Justin and Agrawal, Kumar Krishna and Mo, Xiangxi and Kalra, Sukrit and Schafhalter, Peter and Leong, Eric and Wang, Xin and Balaji, Bharathan and Gonzalez, Joseph E and Stoica, Ion},
  booktitle={Proceedings of the European Conference on Computer Vision (ECCV)},
  year={2022},
}

@inproceedings{schafhalter2023leveraging,
  title={Leveraging cloud computing to make autonomous vehicles safer},
  author={Schafhalter, Peter and Kalra, Sukrit and Xu, Le and Gonzalez, Joseph E and Stoica, Ion},
  booktitle={2023 IEEE/RSJ International Conference on Intelligent Robots and Systems (IROS)},
  pages={5559--5566},
  year={2023},
  organization={IEEE}
}

@inproceedings{ghoshal,
author = {Ghoshal, Moinak and Kong, Z. Jonny and Xu, Qiang and Lu, Zixiao and Aggarwal, Shivang and Khan, Imran and Li, Yuanjie and Hu, Y. Charlie and Koutsonikolas, Dimitrios},
title = {An in-depth study of uplink performance of 5G mmWave networks},
year = {2022},
isbn = {9781450393935},
publisher = {Association for Computing Machinery},
address = {New York, NY, USA},
url = {https://doi.org/10.1145/3538394.3546042},
doi = {10.1145/3538394.3546042},
booktitle = {Proceedings of the ACM SIGCOMM Workshop on 5G and Beyond Network Measurements, Modeling, and Use Cases},
pages = {29–35},
numpages = {7},
location = {Amsterdam, Netherlands},
series = {5G-MeMU '22}
}

@INPROCEEDINGS{razzaghpour,
  author={Razzaghpour, Mohammad and Bockelmann, Carsten and Dekorsy, Armin and Hensel, Jan and Jochim, Wilhelm and Kus, Mehmet},
  booktitle={2024 IEEE Globecom Workshops (GC Wkshps)}, 
  title={Empirical Evaluation of Bit Rate and Latency in a Private 5G Cell for Slow-Speed Vehicles in an Urban Environment}, 
  year={2024},
  volume={},
  number={},
  pages={1-7},
  doi={10.1109/GCWkshp64532.2024.11100938}}

@inproceedings{driving-michigan,
  title = {{The Architectural Implications of Autonomous Driving: Constraints and Acceleration}},
  author = {Lin, Shih-Chieh and Zhang, Yunqi and Hsu, Chang-Hong and Skach, Matt and Haque, Md E. and Tang, Lingjia and Mars, Jason},
  booktitle = {Proceedings of the 23\textsuperscript{rd} International Conference on Architectural Support for Programming Languages and Operating Systems (ASPLOS)},
  year = {2018},
  isbn = {978-1-4503-4911-6},
  location = {Williamsburg, VA, USA},
  pages = {751--766},
  numpages = {16},
  url = {http://doi.acm.org/10.1145/3173162.3173191},
  doi = {10.1145/3173162.3173191},
  acmid = {3173191},
}

@inproceedings{kitti-detection,
  title = {{Are we ready for Autonomous Driving? The KITTI Vision Benchmark Suite}},
  author = {Geiger, Andreas and Lenz, Philip and Urtasun, Raquel},

  booktitle = {Proceedings of the IEEE Conference on Computer Vision and Pattern Recognition (CVPR)},
  year = {2012}
}

@inproceedings{chameleon,
  title = {{Chameleon: Scalable Adaptation of Video Analytics}},
  author = {Jiang, Junchen and Ananthanarayanan, Ganesh and Bodik, Peter and Sen, Siddhartha and Stoica, Ion},
  booktitle = {Proceedings of the ACM Special Interest Group on Data Communication Conference (SIGCOMM)},
  year = {2018},
  isbn = {978-1-4503-5567-4},
  location = {Budapest, Hungary},
  pages = {253--266},
  numpages = {14},
  url = {http://doi.acm.org/10.1145/3230543.3230574},
  doi = {10.1145/3230543.3230574},
  acmid = {3230574},
}

@inproceedings{ren2015faster,
  title={{Faster R-CNN: Towards Real-Time Object Detection with Region Proposal Networks}},
  author={Ren, Shaoqing and He, Kaiming and Girshick, Ross and Sun, Jian},
  booktitle={Proceedings of the International Conferences on Advances in Neural Information Processing Systems (NeurIPS)},
  pages={91--99},
  year={2015}
}

@inproceedings{tan20efficientdet,
  title={{EfficientDet: Scalable and Efficient Object Detection}},
  author={Mingxing Tan and Ruoming Pang and Quoc V. Le},
  booktitle={Proceedings of the IEEE Conference on Computer Vision and Pattern Recognition (CVPR)},
  year={2020}
}

@inproceedings{rhinehart2018r2p2,
  title={{R2P2: A Reparameterized Pushforward Policy for Diverse, Precise Generative Path Forecasting}},
  author={Rhinehart, Nicholas and Kitani, Kris M and Vernaza, Paul},
  booktitle={Proceedings of the European Conference on Computer Vision (ECCV)},
  pages={772--788},
  year={2018}
}

@inproceedings{coco,
  title={{Microsoft CoCo: Common Objects in Context}},
  author={Lin, Tsung-Yi and Maire, Michael and Belongie, Serge and Hays, James and Perona, Pietro and Ramanan, Deva and Doll{\'a}r, Piotr and Zitnick, C Lawrence},
  booktitle={Proceedings of the European Conference on Computer Vision (ECCV)},
  pages={740--755},
  year={2014},
  organization={Springer}
}

@inproceedings{clockwork,
  author = {Gujarati, Arpan and Karimi, Reza and Alzayat, Safya and Kaufmann, Antoine and Vigfusson, Ymir and Mace, Jonathan},
  title = {{Serving DNNs like Clockwork: Performance Predictability from the Bottom Up}},
  booktitle = {Proceedings of the 14\textsuperscript{th} USE\-NIX Symposium on Operating Systems Design and Implementation (OSDI)},
  year = {2020},
  month = nov,
  location = {Banff, Canada},
}

@inproceedings{clipper,
  title = {{Clipper: A Low-Latency Online Prediction Serving System}},
  author = {Crankshaw, Daniel and Wang, Xin and Zhou, Guilio and Franklin, Michael J and Gonzalez, Joseph E and Stoica, Ion},
  booktitle = {Proceedings of the 14\textsuperscript{th} USE\-NIX Conference on Networked Systems Design and Implementation (NSDI)},
  pages = {613--627},
  year = {2017}
}

@inproceedings{streaming-perception,
  author = {Li, Mengtian and Wang, Yuxiong and Ramanan, Deva},
  title = {{Towards Streaming Perception}},
  booktitle={Proceedings of the European Conference on Computer Vision (ECCV)},
  year = {2020},
  month = aug,
  location = {Glasgow, Scotland},
}

@inproceedings{videostorm,
  title = {{Live Video Analytics at Scale with Approximation and Delay-Tolerance}},
  author = {Zhang, Haoyu and Ananthanarayanan, Ganesh and Bodik, Peter and Philipose, Matthai and Bahl, Paramvir and Freedman, Michael J},
  booktitle = {Proceedings of the 14\textsuperscript{th} USE\-NIX Conference on Networked Systems Design and Implementation (NSDI)},
  year = {2017}
}

@inproceedings{caesar_nuscenes_2020,
  title = {{{nuScenes}}: {{A Multimodal Dataset}} for {{Autonomous Driving}}},
  shorttitle = {{{nuScenes}}},
  booktitle = {Proceedings of the IEEE Conference on Computer Vision and Pattern Recognition (CVPR)},
  author = {Caesar, Holger and Bankiti, Varun and Lang, Alex H. and Vora, Sourabh and Liong, Venice Erin and Xu, Qiang and Krishnan, Anush and Pan, Yu and Baldan, Giancarlo and Beijbom, Oscar},
  year = {2020},
  month = jun,
  pages = {11618--11628},
  publisher = {{IEEE}},
  address = {{Seattle, WA, USA}},
  doi = {10.1109/CVPR42600.2020.01164},
  isbn = {978-1-72817-168-5},
  language = {en}
}

@inproceedings{gog2021pylot,
  title = {{Pylot: A Modular Platform for Exploring Latency-Accuracy Tradeoffs in Autonomous Vehicles}},
  author = {Gog, Ionel and Kalra, Sukrit and Schafhalter, Peter and Wright, Matthew A. and Gonzalez, Joseph E. and Stoica, Ion},
  booktitle={Proceedings of the IEEE International Conference on Robotics and Automation (ICRA)},  
  pages = {8806--8813},
  organization = {IEEE},
  year = {2021},
}

@inproceedings{fogros,
  title={FogROS: An Adaptive Framework for Automating Fog Robotics Deployment},
  author={Chen, Kaiyuan Eric and Liang, Yafei and Jha, Nikhil and Ichnowski, Jeffrey and Danielczuk, Michael and Gonzalez, Joseph and Kubiatowicz, John and Goldberg, Ken},
  booktitle={2021 IEEE 17th International Conference on Automation Science and Engineering (CASE)},
  pages={2035--2042},
  year={2021},
  organization={IEEE}
}

@inproceedings{sevilla2022compute,
  title={Compute trends across three eras of machine learning},
  author={Sevilla, Jaime and Heim, Lennart and Ho, Anson and Besiroglu, Tamay and Hobbhahn, Marius and Villalobos, Pablo},
  booktitle={2022 International Joint Conference on Neural Networks (IJCNN)},
  pages={1--8},
  year={2022},
  organization={IEEE}
}

@inproceedings{ghoshal2022depth,
  title={An in-depth study of uplink performance of 5G mmWave networks},
  author={Ghoshal, Moinak and Kong, Z Jonny and Xu, Qiang and Lu, Zixiao and Aggarwal, Shivang and Khan, Imran and Li, Yuanjie and Hu, Y Charlie and Koutsonikolas, Dimitrios},
  booktitle={Proceedings of the ACM SIGCOMM Workshop on 5G and Beyond Network Measurements, Modeling, and Use Cases},
  pages={29--35},
  year={2022}
}

@inproceedings{infaas,
  title={INFaaS: Automated Model-less Inference Serving.},
  author={Romero, Francisco and Li, Qian and Yadwadkar, Neeraja J and Kozyrakis, Christos},
  booktitle={USENIX Annual Technical Conference},
  pages={397--411},
  year={2021}
}

@inproceedings{zhang2021emp,
  title={Emp: Edge-assisted multi-vehicle perception},
  author={Zhang, Xumiao and Zhang, Anlan and Sun, Jiachen and Zhu, Xiao and Guo, Y Ethan and Qian, Feng and Mao, Z Morley},
  booktitle={Proceedings of the 27th Annual International Conference on Mobile Computing and Networking},
  pages={545--558},
  year={2021}
}

@inproceedings{qiu2018avr,
  title={Avr: Augmented vehicular reality},
  author={Qiu, Hang and Ahmad, Fawad and Bai, Fan and Gruteser, Marco and Govindan, Ramesh},
  booktitle={Proceedings of the 16th Annual International Conference on Mobile Systems, Applications, and Services},
  pages={81--95},
  year={2018}
}

@inproceedings{resnet,
  title={Deep residual learning for image recognition},
  author={He, Kaiming and Zhang, Xiangyu and Ren, Shaoqing and Sun, Jian},
  booktitle={Proceedings of the IEEE conference on computer vision and pattern recognition},
  pages={770--778},
  year={2016}
}

@inproceedings{xu2022smartadapt,
  title={SMARTADAPT: Multi-branch Object Detection Framework for Videos on Mobiles},
  author={Xu, Ran and Mu, Fangzhou and Lee, Jayoung and Mukherjee, Preeti and Chaterji, Somali and Bagchi, Saurabh and Li, Yin},
  booktitle={Proceedings of the IEEE/CVF Conference on Computer Vision and Pattern Recognition},
  pages={2528--2538},
  year={2022}
}

@inproceedings{liu2021swin,
  title={Swin transformer: Hierarchical vision transformer using shifted windows},
  author={Liu, Ze and Lin, Yutong and Cao, Yue and Hu, Han and Wei, Yixuan and Zhang, Zheng and Lin, Stephen and Guo, Baining},
  booktitle={Proceedings of the IEEE/CVF international conference on computer vision},
  pages={10012--10022},
  year={2021}
}

@inproceedings{liu2022convnet,
  title={A convnet for the 2020s},
  author={Liu, Zhuang and Mao, Hanzi and Wu, Chao-Yuan and Feichtenhofer, Christoph and Darrell, Trevor and Xie, Saining},
  booktitle={Proceedings of the IEEE/CVF Conference on Computer Vision and Pattern Recognition},
  pages={11976--11986},
  year={2022}
}

@inproceedings{zhai2022scaling,
  title={Scaling vision transformers},
  author={Zhai, Xiaohua and Kolesnikov, Alexander and Houlsby, Neil and Beyer, Lucas},
  booktitle={Proceedings of the IEEE/CVF Conference on Computer Vision and Pattern Recognition},
  pages={12104--12113},
  year={2022}
}

@inproceedings{carion2020end,
  title={End-to-end object detection with transformers},
  author={Carion, Nicolas and Massa, Francisco and Synnaeve, Gabriel and Usunier, Nicolas and Kirillov, Alexander and Zagoruyko, Sergey},
  booktitle={Computer Vision--ECCV 2020: 16th European Conference, Glasgow, UK, August 23--28, 2020, Proceedings, Part I 16},
  year={2020},
  organization={Springer}
}

@inproceedings{zhang2022dino,
  title={Dino: Detr with improved denoising anchor boxes for end-to-end object detection},
  author={Zhang, Hao and Li, Feng and Liu, Shilong and Zhang, Lei and Su, Hang and Zhu, Jun and Ni, Lionel and Shum, Harry},
  booktitle={International Conference on Learning Representations},
  year={2022}
}

@inproceedings{kumar2012cloud,
  title={A cloud-assisted design for autonomous driving},
  author={Kumar, Swarun and Gollakota, Shyamnath and Katabi, Dina},
  booktitle={Proceedings of the first edition of the MCC workshop on Mobile cloud computing},
  pages={41--46},
  year={2012}
}

@article{cui2020offloading,
  title={Offloading autonomous driving services via edge computing},
  author={Cui, Mingyue and Zhong, Shipeng and Li, Boyang and Chen, Xu and Huang, Kai},
  journal={IEEE Internet of Things Journal},
  volume={7},
  number={10},
  pages={10535--10547},
  year={2020},
  publisher={IEEE}
}

@inproceedings{chen2019f,
  title={F-cooper: Feature based cooperative perception for autonomous vehicle edge computing system using 3D point clouds},
  author={Chen, Qi and Ma, Xu and Tang, Sihai and Guo, Jingda and Yang, Qing and Fu, Song},
  booktitle={Proceedings of the 4th ACM/IEEE Symposium on Edge Computing},
  pages={88--100},
  year={2019}
}

@inproceedings{sun2018learning,
  title={Learning-based task offloading for vehicular cloud computing systems},
  author={Sun, Yuxuan and Guo, Xueying and Zhou, Sheng and Jiang, Zhiyuan and Liu, Xin and Niu, Zhisheng},
  booktitle={2018 IEEE International Conference on Communications (ICC)},
  year={2018},
  organization={IEEE}
}

@article{sun2018cooperative,
  title={Cooperative task scheduling for computation offloading in vehicular cloud},
  author={Sun, Fei and Hou, Fen and Cheng, Nan and Wang, Miao and Zhou, Haibo and Gui, Lin and Shen, Xuemin},
  journal={IEEE Transactions on Vehicular Technology},
  volume={67},
  number={11},
  pages={11049--11061},
  year={2018},
  publisher={IEEE}
}

@INPROCEEDINGS{zegura1999utility,
  title     = "Utility max-min: an application-oriented bandwidth allocation
               scheme",
  booktitle = "{IEEE} {INFOCOM} '99. Conference on Computer Communications.
               Proceedings. Eighteenth Annual Joint Conference of the {IEEE}
               Computer and Communications Societies. The Future is Now (Cat.
               {No.99CH36320})",
  author    = "Cao, Zhiruo and Zegura, E W",
  publisher = "IEEE",
  volume    =  2,
  pages     = "793--801 vol.2",
  year      =  1999
}

@INPROCEEDINGS{google2015bwe,
  title     = "{BwE}: Flexible, Hierarchical Bandwidth Allocation for {WAN}
               Distributed Computing",
  booktitle = "Proceedings of the 2015 {ACM} Conference on Special Interest
               Group on Data Communication",
  author    = "Kumar, Alok and Jain, Sushant and Naik, Uday and Raghuraman,
               Anand and Kasinadhuni, Nikhil and Zermeno, Enrique Cauich and
               Gunn, C Stephen and Ai, Jing and Carlin, Bj{\"o}rn and
               Amarandei-Stavila, Mihai and Robin, Mathieu and Siganporia, Aspi
               and Stuart, Stephen and Vahdat, Amin",
  publisher = "Association for Computing Machinery",
  pages     = "1--14",
  series    = "SIGCOMM '15",
  month     =  aug,
  year      =  2015,
  address   = "New York, NY, USA",
  location  = "London, United Kingdom"
}

@INPROCEEDINGS{narayanan20215g-performance,
  title     = "A variegated look at {5G} in the wild: performance, power, and
               {QoE} implications",
  booktitle = "Proceedings of the 2021 {ACM} {SIGCOMM} 2021 Conference",
  author    = "Narayanan, Arvind and Zhang, Xumiao and Zhu, Ruiyang and Hassan,
               Ahmad and Jin, Shuowei and Zhu, Xiao and Zhang, Xiaoxuan and
               Rybkin, Denis and Yang, Zhengxuan and Mao, Zhuoqing Morley and
               Qian, Feng and Zhang, Zhi-Li",
  publisher = "Association for Computing Machinery",
  pages     = "610--625",
  series    = "SIGCOMM '21",
  month     =  aug,
  year      =  2021,
  address   = "New York, NY, USA",
  location  = "Virtual Event, USA"
}

@inproceedings{wang2017cooperative,
  title={Cooperative autonomous driving for traffic congestion avoidance through vehicle-to-vehicle communications},
  author={Wang, Nannan and Wang, Xi and Palacharla, Paparao and Ikeuchi, Tadashi},
  booktitle={2017 IEEE Vehicular Networking Conference (VNC)},
  pages={327--330},
  year={2017},
  organization={IEEE}
}

@inproceedings{waymo-open-motion,
  title={Large scale interactive motion forecasting for autonomous driving: The waymo open motion dataset},
  author={Ettinger, Scott and Cheng, Shuyang and Caine, Benjamin and Liu, Chenxi and Zhao, Hang and Pradhan, Sabeek and Chai, Yuning and Sapp, Ben and Qi, Charles R and Zhou, Yin and others},
  booktitle={Proceedings of the IEEE/CVF International Conference on Computer Vision},
  pages={9710--9719},
  year={2021}
}

@inproceedings{vaswani2017attention,
 author = {Vaswani, Ashish and Shazeer, Noam and Parmar, Niki and Uszkoreit, Jakob and Jones, Llion and Gomez, Aidan N and Kaiser, \L ukasz and Polosukhin, Illia},
 booktitle = {Advances in Neural Information Processing Systems},
 editor = {I. Guyon and U. Von Luxburg and S. Bengio and H. Wallach and R. Fergus and S. Vishwanathan and R. Garnett},
 pages = {},
 publisher = {Curran Associates, Inc.},
 title = {Attention is All you Need},
 url = {https://proceedings.neurips.cc/paper_files/paper/2017/file/3f5ee243547dee91fbd053c1c4a845aa-Paper.pdf},
 volume = {30},
 year = {2017}
}

@inproceedings{xu2020understanding,
  title={Understanding operational 5G: A first measurement study on its coverage, performance and energy consumption},
  author={Xu, Dongzhu and Zhou, Anfu and Zhang, Xinyu and Wang, Guixian and Liu, Xi and An, Congkai and Shi, Yiming and Liu, Liang and Ma, Huadong},
  booktitle={Proceedings of the Annual conference of the ACM Special Interest Group on Data Communication on the applications, technologies, architectures, and protocols for computer communication},
  pages={479--494},
  year={2020}
}

@InProceedings{tan2019efficientnet,
  title = 	 {{E}fficient{N}et: Rethinking Model Scaling for Convolutional Neural Networks},
  author =       {Tan, Mingxing and Le, Quoc},
  booktitle = 	 {Proceedings of the 36th International Conference on Machine Learning},
  pages = 	 {6105--6114},
  year = 	 {2019},
  editor = 	 {Chaudhuri, Kamalika and Salakhutdinov, Ruslan},
  volume = 	 {97},
  series = 	 {Proceedings of Machine Learning Research},
  month = 	 {09--15 Jun},
  publisher =    {PMLR},
  url = 	 {https://proceedings.mlr.press/v97/tan19a.html}
}

@inproceedings{imagenet,
  title={Imagenet: A large-scale hierarchical image database},
  author={Deng, Jia and Dong, Wei and Socher, Richard and Li, Li-Jia and Li, Kai and Fei-Fei, Li},
  booktitle={2009 IEEE conference on computer vision and pattern recognition},
  pages={248--255},
  year={2009},
  organization={Ieee}
}

@inproceedings{bdd-100k,
  title={Bdd100k: A diverse driving dataset for heterogeneous multitask learning},
  author={Yu, Fisher and Chen, Haofeng and Wang, Xin and Xian, Wenqi and Chen, Yingying and Liu, Fangchen and Madhavan, Vashisht and Darrell, Trevor},
  booktitle={Proceedings of the IEEE/CVF conference on computer vision and pattern recognition},
  pages={2636--2645},
  year={2020}
}

@inproceedings{ra2011odessa,
  title={Odessa: enabling interactive perception applications on mobile devices},
  author={Ra, Moo-Ryong and Sheth, Anmol and Mummert, Lily and Pillai, Padmanabhan and Wetherall, David and Govindan, Ramesh},
  booktitle={Proceedings of the 9th international conference on Mobile systems, applications, and services},
  pages={43--56},
  year={2011}
}

@inproceedings{xu2022litereconfig,
  title={Litereconfig: Cost and content aware reconfiguration of video object detection systems for mobile gpus},
  author={Xu, Ran and Lee, Jayoung and Wang, Pengcheng and Bagchi, Saurabh and Li, Yin and Chaterji, Somali},
  booktitle={Proceedings of the Seventeenth European Conference on Computer Systems},
  pages={334--351},
  year={2022}
}

@inproceedings{liu2019edge-assisted-detection,
  title={Edge assisted real-time object detection for mobile augmented reality},
  author={Liu, Luyang and Li, Hongyu and Gruteser, Marco},
  booktitle={The 25th annual international conference on mobile computing and networking},
  pages={1--16},
  year={2019}
}

@inproceedings{ekya,
  title={Ekya: Continuous learning of video analytics models on edge compute servers},
  author={Bhardwaj, Romil and Xia, Zhengxu and Ananthanarayanan, Ganesh and Jiang, Junchen and Shu, Yuanchao and Karianakis, Nikolaos and Hsieh, Kevin and Bahl, Paramvir and Stoica, Ion},
  booktitle={19th USENIX Symposium on Networked Systems Design and Implementation (NSDI 22)},
  pages={119--135},
  year={2022}
}

@inproceedings{kang2017noscope,
  author = {Kang, Daniel and Emmons, John and Abuzaid, Firas and Bailis, Peter and Zaharia, Matei},
  title = {NoScope: optimizing neural network queries over video at scale},
  year = {2017},
  issue_date = {August 2017},
  publisher = {VLDB Endowment},
  volume = {10},
  number = {11},
  issn = {2150-8097},
  url = {https://doi.org/10.14778/3137628.3137664},
  doi = {10.14778/3137628.3137664},
  journal = {Proc. VLDB Endow.},
  month = {aug},
  pages = {1586–1597},
  numpages = {12}
}

@inproceedings{li2020reducto,
  title={Reducto: On-camera filtering for resource-efficient real-time video analytics},
  author={Li, Yuanqi and Padmanabhan, Arthi and Zhao, Pengzhan and Wang, Yufei and Xu, Guoqing Harry and Netravali, Ravi},
  booktitle={Proceedings of the Annual conference of the ACM Special Interest Group on Data Communication on the applications, technologies, architectures, and protocols for computer communication},
  pages={359--376},
  year={2020}
}

@inproceedings{du2020server,
  title={Server-driven video streaming for deep learning inference},
  author={Du, Kuntai and Pervaiz, Ahsan and Yuan, Xin and Chowdhery, Aakanksha and Zhang, Qizheng and Hoffmann, Henry and Jiang, Junchen},
  booktitle={Proceedings of the Annual conference of the ACM Special Interest Group on Data Communication on the applications, technologies, architectures, and protocols for computer communication},
  pages={557--570},
  year={2020}
}

@inproceedings{nagaraj2016-numfabric,
author = {Nagaraj, Kanthi and Bharadia, Dinesh and Mao, Hongzi and Chinchali, Sandeep and Alizadeh, Mohammad and Katti, Sachin},
title = {NUMFabric: Fast and Flexible Bandwidth Allocation in Datacenters},
year = {2016},
isbn = {9781450341936},
publisher = {Association for Computing Machinery},
address = {New York, NY, USA},
url = {https://doi.org/10.1145/2934872.2934890},
doi = {10.1145/2934872.2934890},
booktitle = {Proceedings of the 2016 ACM SIGCOMM Conference},
pages = {188–201},
numpages = {14},
location = {Florianopolis, Brazil},
series = {SIGCOMM '16}
}

@inproceedings{zong2023detrs,
  title={Detrs with collaborative hybrid assignments training},
  author={Zong, Zhuofan and Song, Guanglu and Liu, Yu},
  booktitle={Proceedings of the IEEE/CVF international conference on computer vision},
  pages={6748--6758},
  year={2023}
}

@TECHREPORT{nhtsa2023recalls,
  title       = "{NHTSA} 2022 Annual Report Safety Recalls",
  author      = "{National Highway Traffic Safety Administration}",
  institution = "National Highway Traffic Safety Administration",
  month       =  mar,
  year        =  2023
}

@MISC{cable2023averagecellularcost,
  title        = "Worldwide Mobile Data Pricing 2023",
  booktitle    = "Cable.co.uk",
  author       = "Howdle, Dan",
  howpublished = "\url{https://www.cable.co.uk/mobiles/worldwide-data-pricing/}",
  note         = "Accessed: 2024-6-13",
  language     = "en"
}

@TECHREPORT{5gdeploymentspecs,
  title       = "Report {ITU-R} M.2410-0: Minimum Requirements related to
                 Technical Performance for {IMT-2020} Radio Interface(s)",
  author      = "{International Telecommunications Union}",
  number      = "M.2410-0",
  institution = "International Telecommunications Union",
  year        =  2017
}

@TECHREPORT{mckinsey2024avleaders,
  title       = "Autonomous Vehicles Moving Forward: Perspectives from Industry
                 Leaders",
  author      = "{Derek Chiao, Johannes Deichmann, Kersten Heineke, Ani Kelkar,
                 Martin Kellner, Elizabeth Scarinci, Dmitry Tolstinev}",
  institution = "McKinsey",
  month       =  jan,
  year        =  2024
}

@TECHREPORT{trafficsafety2023drivingmin,
  title       = "American Driving Survey: 2022",
  author      = "Rebecca Steinbach, Brian Tefft",
  institution = "AAA Foundation for Traffic Safety",
  month       =  sep,
  year        =  2023
}

@MISC{bureauoftransportation2024age,
  title    = "Average Age of Automobiles and Trucks in Operation in the United
              States",
  author   = "{Bureau of Transportation Statistics}",
  month    =  jan,
  year     =  2024
}

@misc{nhtsa-cruise-incident-2023,
  title={Part 573 Safety Recall Report 23E-086},
  author={National Highway Traffic Safety Administration},
  year=2023,
  month=nov,
  date=7,
}

@MISC{ericsson-5gcoverage,
  title        = "Network coverage forecast – Ericsson Mobility Report",
  author = {Ericsson},
  booktitle    = "ericsson.com",
  month        =  nov,
  year         =  2020,
  howpublished = "\url{https://www.ericsson.com/en/reports-and-papers/mobility-report/dataforecasts/network-coverage}",
  note         = "Accessed: 2024-9-19",
  language     = "en"
}

@misc{rfc9000,
    series =    {Request for Comments},
    number =    9000,
    howpublished =  {RFC 9000},
    publisher = {RFC Editor},
    doi =       {10.17487/RFC9000},
    url =       {https://www.rfc-editor.org/info/rfc9000},
    author =    {Jana Iyengar and Martin Thomson},
    title =     {{QUIC: A UDP-Based Multiplexed and Secure Transport}},
    pagetotal = 151,
    year =      2021,
    month =     may,
}

@misc{cruise-roscon,
  author = {Nicolo Valigi},
  title = {{Lessons Learned Building a Self-Driving Car on ROS}},
  howpublished = {\url{https://roscon.ros.org/2018/presentations/ROSCon2018_LessonsLearnedSelfDriving.pdf}},
  year = {2018},
}

@misc{waymo-safety-report,
  author = {{Waymo}},
  title = {{Waymo Safety Report: On the Road to Fully Self-Driving}},
  howpublished = {\url{https://storage.googleapis.com/sdc-prod/v1/safety-report/SafetyReport2018.pdf}}
}

@misc{kpmg-self-driving-report,
  author = {{KPMG}},
  title = {{Self-driving cars: The next revolution}},
  note = {\url{https://institutes.kpmg.us/content/dam/institutes/en/manufacturing/pdfs/2017/self-driving-cars-next-revolution-new.pdf}}
}

@misc{mckinsey-50mins,
  author = {{Michele Bertoncello, and Dominik Wee}},
  title = {{Ten Ways Autonomous Driving Could Redefine the Automotive World}},
  howpublished = {\url{https://tinyurl.com/2srpyv8d}}
}

@misc{apollo-baidu,
  author = {{Baidu}},
  title = {{Apollo 3.0 Software Architecture}},
  howpublished = {\url{https://tinyurl.com/mhd6dfka}}
}

@misc{argo-data,
  title = {Argoverse},
  author = {Argoverse},
  howpublished = {\url{https://www.argoverse.org/}},
}

@misc{nhtsa-sae-automation,
  author = {National Highway Traffic Safety Administration},
  title = {{Automated Vehicles for Safety}},
  howpublished = {\url{https://www.nhtsa.gov/technology-innovation/automated-vehicles-safety}},
}

@misc{waymo-av-sensors,
  author = {Waymo},
  title = {{Introducing the 5\textsuperscript{th} Generation Waymo Driver}},
  howpublished = {\url{https://blog.waymo.com/2020/03/introducing-5th-generation-waymo-driver.html}},
}

@misc{cruise-austin,
  title = {Cruise to launch robotaxi services in Austin, Phoenix before end of 2022},
author = "Cruise",
  howpublished = {\url{https://techcrunch.com/2022/09/12/cruise-to-launch-robotaxi-services-in-austin-phoenix-before-end-of-2022/}},
}

@misc{waymo-los-angeles,
  title = {Next Stop for Waymo One: Los Angeles},
  author = {Waymo},
  howpublished = {\url{https://waymo.com/blog/2022/10/next-stop-for-waymo-one-los-angeles.html}},
}

@misc{compound-ai-systems-blog,
  title        = "The Shift from Models to Compound {AI} Systems",
  author       = "Seita, Daniel and {Matei Zaharia, Omar Khattab, Lingjiao
                  Chen, Jared Quincy Davis, Heather Miller, Chris Potts, James
                  Zou, Michael Carbin, Jonathan Frankle, Naveen Rao, Ali
                  Ghodsi}",
  howpublished = "\url{https://bair.berkeley.edu/blog/2024/02/18/compound-ai-systems/}",
  note         = "Accessed: 2024-2-20"
}

@misc{waymo-scaling-to-four-cities,
  author = "Waymo",
  title = {Scaling Waymo One safely across four cities this year},
  howpublished = {\url{https://waymo.com/blog/2024/03/scaling-waymo-one-safely-across-four-cities-this-year/}},
  month = mar,
  date = 13,
  year = 2024
}

@misc{ntsa-uber-collision,
    author = {National Highway Traffic Safety Administration},
    title = "Collision Between Vehicle Controlled by Developmental Automated Driving System and Pedestrian",
    month = "mar",
    day = 18,
    year = 2018,
    howpublished = {\url{https://www.ntsb.gov/investigations/accidentreports/reports/har1903.pdf}}
}

@misc{bloomberg-self-driving-is-going-nowhere,
  author = {Chafkin, Max},
  title = {Even After \$100 Billion, Self-Driving Cars Are Going Nowhere},
  day = 5,
  month = oct,
  year = 2022,
  howpublished = {\url{https://www.bloomberg.com/news/features/2022-10-06/even-after-100-billion-self-driving-cars-are-going-nowhere}}
}

@misc{2024-h100-price,
  author = {Shilov, Anton},
  title = {Nvidia's H100 AI GPUs cost up to four times more than AMD's competing MI300X — AMD's chips cost \$10 to \$15K apiece; Nvidia's H100 has peaked beyond \$40,000: Report},
  day = 2,
  month = feb,
  year = 2024,
  howpublished = {\url{https://www.tomshardware.com/tech-industry/artificial-intelligence/nvidias-h100-ai-gpus-cost-up-to-four-times-more-than-amds-competing-mi300x-amds-chips-cost-dollar10-to-dollar15k-apiece-nvidias-h100-has-peaked-beyond-dollar40000}}
}

@misc{tesla-model-3-price,
  author = {Brandt, Eric},
  title = {2024 Tesla Model 3},
  day = 8,
  month = mar,
  year = 2024,
  howpublished = {\url{https://www.kbb.com/tesla/model-3/}},
  note = {Accessed 2024-6-11}
}

@misc{drive-orin-spec,
  title = {DRIVE AGX Orin Developer Kit}, 
  author = "NVIDIA",
  howpublished = {\url{https://developer.nvidia.com/drive/agx}},
  note = {Accessed 2024-6-11}
}

@misc{h100-spec,
  title = {NVIDIA H100 Tensor Core GPU},
  author = "NVIDIA",
  howpublished = {\url{https://www.nvidia.com/en-us/data-center/h100/}},
  note = {Accessed 2024-6-11}
}

@misc{waymo-fleet-response,
  title = {Fleet response: Lending a helpful hand to Waymo’s autonomously driven vehicles},
  author = {The Waymo Team},
  day = 21,
  month = may,
  year = 2024,
  howpublished = {\url{https://waymo.com/blog/2024/05/fleet-response/}},
}

@misc{siemens-av-data,
  title = {The Data Deluge: What do we do with the data generated by AVs?},
  author = {Götz, Florian},
  day = 22,
  month = jan,
  year = 2021,
  howpublished = {\url{https://blogs.sw.siemens.com/polarion/the-data-deluge-what-do-we-do-with-the-data-generated-by-avs/}},
}

@MISC{zipit2023wirelesscost,
  title        = "How to {Cost-Effectively} Manage {IoT} Data Plans",
  author       = "Heredia, Ralph",
  publisher    = "Zipit",
  month        =  jun,
  year         =  2023,
  howpublished = "\url{https://www.zipitwireless.com/blog/how-to-cost-effectively-manage-iot-data-plans}",
  note         = "Accessed: 2024-6-15",
  language     = "en"
}

@misc{karpathy-keynote-cvpr-wad-2021,
  title = {[CVPR'21 WAD] Keynote - Andrej Karpathy, Tesla},
  author = {Karpathy, Andrej},
  day = 20,
  month = jun,
  year = 2021,
  howpublished = {\url{https://youtu.be/g6bOwQdCJrc}},
}

@misc{nvidia-localization,
  title = {DRIVE Labs: How Localization Helps Vehicles Find Their Way},
  author = {Alarcon, Nefi},
  day = 20,
  month = jan,
  year = 2020,
  howpublished = {\url{https://developer.nvidia.com/blog/drive-labs-how-localization-helps-vehicles-find-their-way/}},
}

@misc{nvidia-drive-faq,
  title = {NVIDIA DRIVE DEVELOPER FAQ},
  author = {NVIDIA},
  howpublished={\url{https://developer.nvidia.com/drive/faq}},
  note = {Accessed 2024-6-23},
}

@misc{lambda-labs-gpu-pricing,
  title = {GPU Cloud},
  author = {Lambda},
  howpublished = {\url{https://lambdalabs.com/service/gpu-cloud}},
  note = {Accessed 2024-6-23},
}

@misc{uber-nyc-pickups,
  title = {Uber Pickups in New York City},
  author = {FiveThirtyEight},
  howpublished = {\url{https://www.kaggle.com/datasets/fivethirtyeight/uber-pickups-in-new-york-city}},
  note = {Accessed 2024-6-23},
}

@MISC{pyimagesearch2022cocomap,
  title        = "Mean Average Precision ({mAP}) Using the {COCO} Evaluator",
  booktitle    = "{PyImageSearch}",
  author       = "Sharma, Aditya",
  month        =  may,
  year         =  2022,
  howpublished = "\url{https://pyimagesearch.com/2022/05/02/mean-average-precision-map-using-the-coco-evaluator/}",
  note         = "Accessed: 2024-6-25"
}

@misc{waymo-6th-gen-driver,
  title = {Meet the 6th-generation Waymo Driver: Optimized for costs, designed to handle more weather, and coming to riders faster than before},
  author = {Jeyachandran, Satish},
  day = 19,
  month = aug,
  year = 2024,
  howpublished = {\url{https://waymo.com/blog/2024/08/meet-the-6th-generation-waymo-driver/}},
  note = "Accessed: 2024-8-20",
}

@misc{waymo-motion-prediction-challenge,
  title = {Motion Prediction},
  author = {Waymo},
  howpublished = {\url{https://waymo.com/open/challenges/2024/motion-prediction/}},
  note = "Accessed: 2024-8-20",
}

@misc{huggingface-object-detection-leaderboard,
  title = {Hugging Face Object Detection Leaderboard},
  author = {Padilla, Rafael and Roberts, Amy},
  day = 18,
  month = sep,
  year = 2023,
  howpublished = {\url{https://huggingface.co/blog/object-detection-leaderboard}},
  note = {Accessed: 2024-8-22},
}

@misc{apollo-auto-github,
  title = {Apollo},
  author = {Apollo},
  howpublished = {\url{https://github.com/ApolloAuto/apollo/}},
  note = {Accessed: 2024-9-5},
}

@misc{effdet-github,
  title = {efficientdet-pytorch},
  author = {Wightman, Ross},
  howpublished = {\url{https://github.com/rwightman/efficientdet-pytorch}},
  note = {Accessed: 2024-9-6},
}

@misc{cbc-solver-2.10.12,
  author       = {John Forrest and
                  Ted Ralphs and
                  Stefan Vigerske and
                  Haroldo Gambini Santos and
                  John Forrest and
                  Lou Hafer and
                  Bjarni Kristjansson and
                  jpfasano and
                  EdwinStraver and
                  Jan-Willem and
                  Miles Lubin and
                  rlougee and
                  a-andre and
                  jpgoncal1 and
                  Samuel Brito and
                  h-i-gassmann and
                  Cristina and
                  Matthew Saltzman and
                  tosttost and
                  Bruno Pitrus and
                  Fumiaki MATSUSHIMA and
                  Patrick Vossler and
                  Ron @ SWGY and
                  to-st},
  title        = {coin-or/Cbc: Release releases/2.10.12},
  month        = aug,
  year         = 2024,
  publisher    = {Zenodo},
  version      = {releases/2.10.12},
  doi          = {10.5281/zenodo.13347261},
  url          = {https://doi.org/10.5281/zenodo.13347261}
}

@misc{pulp-modeler,
  title={PuLP},
  author={Roy, J.S. and Mitchell, Stuart A. and Duquesne, Christophe-Marie and Peschiera, Franco and Phillips Antony},
  howpublished={\url{https://github.com/coin-or/pulp}},
  note={Accessed 2024-9-14},
}

@misc{h100-price-june-2024,
  title={Intel's Gaudi 3 will cost half the price of Nvidia's H100},
  author={Shilov, Anton},
  year=2024,
  month=jun,
  date=6,
  howpublished={\url{https://www.tomshardware.com/pc-components/cpus/intels-gaudi-3-will-cost-half-the-price-of-nvidias-h100}},
}

@misc{drive-orin-usage,
  title={Volvo Cars, Zoox, SAIC and More Join Growing Range of Autonomous Vehicle Makers Using New NVIDIA DRIVE Solutions},
  author={Labrie, Marie},
  year=2021,
  month=apr,
  day=12,
  howpublished={\url{https://nvidianews.nvidia.com/news/volvo-cars-zoox-saic-and-more-join-growing-range-of-autonomous-vehicle-makers-using-new-nvidia-drive-solutions}},
}

@misc{usa-vehicle-registrations-2022,
  title={Table MV-1 - Highway Statistics 2022},
  author={Federal Highway Administration},
  year=2023,
  month=nov,
  howpublished={\url{https://www.fhwa.dot.gov/policyinformation/statistics/2022/mv1.cfm}},
}

@misc{bernstein-ev-revolution-2021,
  author = {Ellinghorst, Arndt and Becker, Meike and Beveridge, Neil and Brackett, Bob  and Chigumira, Danielle and Clint, Oswald and Dillard, Chad and Foran, Brian and Garre, Venugopal and Huang, Jay and Leung, Cherry and Li, Mark and Ma, Zhihan and Sacconaghi, A.M. (Toni) and Salisbury, Jean Ann and Venkateswaran, Deepa and Wang, Lu and Wildhack, Robert and Zechmann, Gunther},
  title = {Electric Revolution 2021: From Dream to Scare to Reality?},
  institution = {Bernstein Autonomous},
  month = aug,
  year = {2021}
}

@misc{ca-dmv-disengagement-reports,
    author={State of California Department of Motor Vehicles},
    title={Disengagement Reports},
    howpublished={\url{https://www.dmv.ca.gov/portal/vehicle-industry-services/autonomous-vehicles/disengagement-reports/}}
}

@misc{waymo-safety-impact,
    title={Waymo Safety Impact},
    author = {Waymo},
    howpublished={\url{https://waymo.com/safety/impact/}},
    note={Accessed 2024-9-18}
}

@misc{waymo-outperforms-humans,
    title={Waymo significantly outperforms comparable human benchmarks over 7+ million miles of rider-only driving},
    author={The Waymo Team},
    year=2023,
    month=dec,
    day=20,
    howpublished={\url{https://waymo.com/blog/2023/12/waymo-significantly-outperforms-comparable-human-benchmarks-over-7-million/}},
}

@MISC{kiwibots,
  title        = "Delivery Robots for Everyone!",
  author = {Kiwibot},
  howpublished = "\url{https://www.kiwibot.com/}",
  note         = "Accessed: 2024-9-18",
  language     = "en"
}

@MISC{zoox,
  title        = "Zoox",
  author = {Zoox},
  howpublished = "\url{https://zoox.com/}",
  note         = "Accessed: 2024-9-18"
}

@misc{autoware-concepts,
  title={Autoware concepts},
author = {Autoware},
  howpublished={\url{https://autowarefoundation.github.io/autoware-documentation/galactic/design/autoware-concepts/}},
  note={Accessed 2024-9-18},
}

@misc{nuttcp,
  author = {Fink, Bill and Scott, Rob},
  title = {nuttcp},
  url = {http://nuttcp.net/},
  version = {8.1.4},
  date = {2016},
}

@misc{iperf3,
  author = {{ESnet}},
  title = {iperf3},
  organization = {Lawrence Berkeley National Laboratory},
  url = {https://github.com/esnet/iperf},
  year = {2024},
}

@MISC{microsoft-quantum,
  title        = "What is Quantum Computing",
  author = {Microsoft},
  howpublished = "\url{https://azure.microsoft.com/en-us/resources/cloud-computing-dictionary/what-is-quantum-computing}",
  note         = "Accessed: 2024-9-19",
  language     = "en"
}

@article{kelly1998ratecontrol,
author = {F P Kelly, A K Maulloo and D K H Tan},
title = {Rate control for communication networks: shadow prices, proportional fairness and stability},
journal = {Journal of the Operational Research Society},
volume = {49},
number = {3},
pages = {237--252},
year = {1998},
publisher = {Taylor \& Francis},
doi = {10.1057/palgrave.jors.2600523},
URL = { 
    
        https://doi.org/10.1057/palgrave.jors.2600523
},
eprint = { 
    
        https://doi.org/10.1057/palgrave.jors.2600523 
}
}

@misc{nuro-food-delivery,
    title={Driverless food delivery: Nuro teams up with Uber Eats to deploy autonomous vehicles in Mountain View},
    author={Gemmet, Andrea and Martin, Malea},
    day=11,
    month=sep,
    year=2022,
    howpublished={\url{https://www.paloaltoonline.com/news/2022/09/11/driverless-food-delivery-nuro-teams-up-with-uber-eats-to-deploy-autonomous-vehicles-in-mountain-view/}}
}

@misc{awss2n-quic_2025,
	title = {aws/s2n-quic},
    author = {Amazon Web Services},
	rights = {Apache-2.0},
	url = {https://github.com/aws/s2n-quic},
	publisher = {Amazon Web Services},
	urldate = {2025-08-29},
	date = {2025-08-28},
	note = {original-date: 2020-06-25T18:27:25Z},
}

@MISC{tesla-robotaxi,
  title        = "Robotaxi",
  author = "Tesla",
  booktitle    = "Tesla",
  howpublished = "\url{https://www.tesla.com/robotaxi}",
  note         = "Accessed: 2025-9-29",
  language     = "en"
}
% {
%   \bibliographystyle{plain}
%   \IfFileExists{bibliography/articles.bib}{
%     % Overleaf
%       \bibliography{bibliography/articles,bibliography/books,bibliography/mine,bibliography/papers,bibliography/patents,bibliography/standards,bibliography/techreports,bibliography/theses,bibliography/urls}
%   }{
%     % Local
%       \bibliography{../bibliography/articles,../bibliography/books,../bibliography/mine,../bibliography/papers,../bibliography/patents,../bibliography/standards,../bibliography/techreports,../bibliography/theses,../bibliography/urls}
%   }
% }
\clearpage
\appendix

\section{Supplementary Material}

\subsection{Economic Feasibility}
\label{s:appendix-feasibility}

In this section, we analyze whether our approach is feasible when taking into account the cost of network transmission (\cref{s:appendix-feasibility-cost}) and compute (\cref{s:appendix-compute-cost}).

\subsubsection{Network Cost}
\label{s:appendix-feasibility-cost}
Commercial cellular network usage is charged primarily by the GB~\cite{zipit2023wirelesscost}. We conduct an analysis of consumer-marketed cellular data plans reported in the Cable.co.uk global mobile data pricing dataset~\cite{cable2023averagecellularcost}. \cref{t:network-costs} shows the cheapest \textit{consumer}-facing (\ie SIM
card) cost per GB of data in a selection of countries, along with the
computed cost per hour of streaming an average of 100 Mbps of data continuously.
We note that we expect wholesale pricing, especially geofenced to a
particular region, to be considerably cheaper.

We see that prices vary widely from as low as \$0.001/GB in Israel, to \$0.75/GB in the US, up to over \$2 in Norway.
This wide range in pricing requires careful consideration in deployment: in countries such as Israel, the price of cellular data transmission running our method is trivial at \$0.04 per hour of driving, assuming an average utilization of 100 Mbps.
In some other countries, including the U.S. with a price of \$33.76 per hour, prices are considerably higher and present an economic obstacle at present to
using remote resources. However, we note that at or below the 10th percentile of global prices – which includes major markets such as India, Italy, and China – mobile networks are cost-effective at \$2.78 per hour of driving. We expect much of the rest of the world to follow to these prices, as median price per GB has continuously decreased $4\times$ over the years our dataset covers, 2019-2024, from \$5.25 to \$1.28. 

In the short-term, in countries with high cellular data prices, operators may choose to reduce
costs by selectively utilizing remote resources to aid in high-stress
driving environments \eg during poor visibility due to weather and busy
urban areas. Alternatively, it would be feasible for AV fleet operators to deploy their own dedicated locale-specific wireless network at cheaper cost.

\subsubsection{Compute Cost}
\label{s:appendix-compute-cost}
Cloud providers offer competitive access to GPUs: Lambda Labs hourly pricing ranges from \$0.80 for an NVIDIA A6000 GPU to \$2.49 for an NVIDIA H100 GPU~\cite{lambda-labs-gpu-pricing}.

Cloud compute offers a number of additional valuable advantages not available on the car. Cloud access allows operators to configure which compute resources to
use based on compute requirements and cost sensitivity.
Remote resources cannot be stolen or damaged in an accident. AV fleet operators can take advantage of statistical multiplexing to share a smaller set of compute for their fleet~\cref{s:appendix:cloud-multiplexing}. Furthermore, model serving systems can optimize resource utilization by batching and scheduling requests~\cite{clipper,infaas,clockwork}, resulting
in further price improvements.

\vspace{1em}
\myparagraph{Total cost.}
We estimate total hourly cost of remote resources at
\$5.27, with \$2.78 from the network (at the 10th global percentile)
and \$2.49 from compute for an H100.
We emphasize that the true cost of cloud compute is likely
lower due to better efficiency when operating at scale.

\begin{table}
\centering
\small
  \begin{tabular}{| c  |c  |c  |c  |c  |}
\hline
\textbf{Rank} & \textbf{Country} & \textbf{\$/GB} & \textbf{\$/Hour}\\
\hline
1 & Singapore & \$0.07 & \$3.30 \\
\hline
2 & Netherlands& \$0.36 & \$16.08 \\
\hline
3 & Norway & \$2.09 & \$94.14 \\
\hline
4 & United States& \$0.75 & \$33.76 \\
\hline
5 & Finland & \$0.26 & \$11.62 \\
\hline
-- & China & \$0.27 & \$12.28 \\
\hline
-- & Israel & \$0.001 & \$0.04 \\
\hline
-- & \textit{10th pct} & \$0.062 & \$2.78 \\
\hline
-- & \textit{Median} & \$0.37 & \$16.84 \\
\hline
\end{tabular}

  \vspace{0.5em}
  \caption{Network costs ranked highest by AV readiness
  score~\cite{kpmg-self-driving-report}. We include China as a major
  AV market~\cite{mckinsey2024avleaders}, Israel as the cheapest cellular
  market, and the 10th percentile and median global country by network
  price. Hourly rates assume an average constant network utilization of 100 Mbps.
  }
\label{t:network-costs}
\end{table}

\subsubsection{Cloud Compute Multiplexing}
\label{s:appendix:cloud-multiplexing}
Sharing upgraded compute costs across a fleet of vehicles presents a significant
opportunity to further reduce costs over upgrading on-vehicle compute due to statistical multiplexing of cloud resources.
Though cars still need to retain their own GPUs, shared upgrades to cloud resources
can instantly benefit fleets of AVs.
% compute upgrades going forward can use multiplexed cloud resources. 
% compared to installing dedicated compute
% hardware in each car. 
The average driver in the U.S. drives only 60.2 minutes per
day~\cite{trafficsafety2023drivingmin}, \ie a vehicle utilization of
$4.2$\%.
This under-utilization is more pronounced for personal vehicles than
autonomous ride-hailing services, which we estimate to be
${\sim}59$\% based on the ratio of peak to average hourly Uber rides in New
York City~\cite{uber-nyc-pickups}.
Considering this under-utilization, the cost of purchasing a single H100 GPU
(${\sim}\$40$k~\cite{2024-h100-price}) is equivalent to renting an H100 in
the cloud for an 44 years for the average American driver, and 3 years for
the average autonomous ride-hailing vehicle.

\subsection{Practical Necessity of \sysname}
% We acknowledge that commercial AVs can achieve safety with on-car resources today.
% In our work, we do not claim that commercial AVs cannot be safe with on-car resources today \sr{double negative}; we have an existence proof to the contrary in 
% This is evidenced by the successful commercial rollouts of Waymo, Cruise, and others across the country~\cite{waymo-los-angeles,cruise-austin,waymo-scaling-to-four-cities}.
We acknowledge that there are limited deployments of commercial AVs today~\cite{waymo-los-angeles,cruise-austin,waymo-scaling-to-four-cities}
which outperform humans on safety benchmarks~\cite{waymo-safety-impact,waymo-safety-report},
and rely on this capability to provide a fallback when network connectivity is unavailable.
% In fact, we rely on this capability to provide a reasonable fallback when network connectivity is unavailable.
However, ``safety'' as both a concept and a metric is continuous, measured as a rate of incident occurrence~\cite{ca-dmv-disengagement-reports,waymo-safety-impact,waymo-safety-report},
and AVs must merely exceed human-level safety for deployment~\cite{waymo-outperforms-humans}.
% and AV deployment has been gated by AVs reaching and surpassing the safety rate of human driver~\cite{waymo-outperforms-humans}.
\sysname seeks to improve safety beyond what on-car systems can provide today and presents an opt-in solution for AV providers to improve the accuracy of their services by leverage cloud and network infrastructure.
% \peter{Softened this claim from ``improves'' to ``seeks to improve''}
% Different AV providers can opt in depending on their priorities.
% While exact widescale deployment requirements are subject to debate~\cite{cmu2017avdeploy}, it is largely accepted that the threshold for AVs to surpass is the safety rate of human driving~\cite{waymo-outperforms-humans}. There is still both benefit and good reason to opportunistically increase safety \textit{beyond} this threshold.\sr{Feel a bit like the above text is overcomplicating a simple message: we improve safety beyond what on-car solutions provide. And different AV providers can opt in depending on their priorities.} 

For AVs with different sets of economic and technical requirements (\eg operating in low-stakes environments and with little compute, such as autonomous delivery robots~\cite{kiwibots,zoox,nuro-food-delivery}), \sysname may provide a viable method of improving existing or expanding functionality such as safe high-speed operation or decreasing the rate of human intervention~\cite{waymo-fleet-response}.

As AVs deployments expand, we expect to see more ``outdated'' AV models on the road\footnote{The average lightweight vehicle age in the U.S. is 12.5 years~\cite{bureauoftransportation2024age}.}.
As SOTA compute hardware performance, and correspondingly SOTA model sizes and requirements, continues to rapidly increase~\cite{sevilla2022compute,liu2022convnet,vit}, outdated hardware will prevent older vehicles from utilizing the latest model advancements.
While upgrades to on-vehicle compute hardware are possible,
they may be difficult to roll out in practice.
As a reference,
fix rates for recalls are only 52-64\%~\cite{nhtsa2023recalls}
despite being free, mandatory upgrades that mitigate critical safety risks (\eg faulty brake systems~\cite{honda2023recall}).
\sysname provides access to SOTA cloud hardware, enabling older AVs to use highly accurate, SOTA models.
% While upgrades to on-vehicle compute hardware are possible, in practice it is very difficult; as a reference, rates for recall service completion – free, mandatory upgrades that fix critical safety risks such as faulty brake systems~\cite{honda2023recall} – are just 52-64\%~\cite{nhtsa2023recalls}.
% \sysname can make it possible for such older AVs to access and use such models.

% prevent an increasing number of on-road vehicles from utilizing the latest advancements in models that could further improve safety.
% While upgrading the on-car compute hardware is possible in theory, in practice it is very difficult; as a reference, rates for recall service completion – free, mandatory upgrades that fix critical safety risks such as faulty brake systems~\cite{honda2023recall} – are just 52-64\%~\cite{nhtsa2023recalls} as many car owners do not service their vehicles due to service duration, required use, and distance from servicing~\cite{cmt-recall-challenge}. \sysname can make it possible for such older AVs to access and use such models.
% \sr{can shrink this para if you need to save space: just need to say that SAVE-B can also alleviate issues with older AVs that might be slow to upgrade their on-car compute, which evidence suggests is often the case even with necessary fixes (cite).}

% Finally, compute hardware is increasingly demanding in its requirements,
Finally, SOTA hardware requires increasingly stringent operating environments,
ranging from high power and cooling needs for GPUs~\cite{driving-michigan}, to strong intolerance to movement or temperature changes for quantum computing~\cite{microsoft-quantum}.
As a result, the only way to integrate methods that require such hardware \textit{is} via the network, and \sysname presents a system that can manage this integration.
Hence, we firmly believe that such a system is useful today and will become more applicable to the real world as current technical and economic trends continue.

\end{document}